\documentclass[review]{fcs}
\usepackage[T1]{fontenc}   
\usepackage[utf8]{inputenc} 
\usepackage{amsmath,amsfonts}
\usepackage{array}
\usepackage{multirow}
\usepackage{textcomp}
\usepackage{stfloats}
\usepackage{url}
\usepackage{verbatim}
\usepackage{graphicx}
\usepackage{cite}
\usepackage{booktabs}
\usepackage{makecell}
\usepackage{color}
\usepackage{tabularray}
\usepackage{CJKutf8}
\usepackage{ragged2e}
\usepackage{tabu}
\usepackage{arydshln}
\usepackage{colortbl}
\usepackage{hyperref}
\usepackage{pifont}

\title{A Survey on Multilingual Large Language Models: Corpora, Alignment, and Bias}
\author*{Yuemei XU}

\author{Ling HU}
\author{Jiayi ZHAO}
\author{Zihan QIU}
\author{Kexin XU}
\author{Yuqi YE}
\author{Hanwen GU}
\address{School of Information Science and Technology, Beijing Foreign Studies University, Beijing 100089, China}
\fcssetup{
  received       = {June 07, 2024},
  accepted       = {November 28, 2024},
  corr-email     = {xuyuemei@bfsu.edu.cn. 
  This work was supported by the National Social Science Foundation (No.24CYY107).},
}

\begin{abstract}
\begin{sloppypar}

Based on the foundation of Large Language Models (LLMs), 
Multilingual LLMs (MLLMs) have been developed to 
address the challenges faced in multilingual natural language processing, 
hoping to achieve knowledge transfer from high-resource languages to low-resource languages. 
However, 
significant limitations and challenges still exist, 
such as language imbalance, multilingual alignment, and inherent bias. 
In this paper, we aim to provide a comprehensive analysis of MLLMs, delving deeply into
discussions surrounding these critical issues.
First of all, we start by presenting an overview of MLLMs, 
covering their evolutions, key techniques, and multilingual capacities. 
Secondly, we explore the multilingual training corpora of MLLMs and 
the multilingual datasets oriented for downstream tasks that are crucial to enhance the cross-lingual capability of MLLMs. 
Thirdly, we survey the state-of-the-art studies of multilingual representations and 
investigate whether the current MLLMs can learn a universal language representation. 
Fourthly, we discuss bias on MLLMs, including its categories, evaluation metrics, and debiasing techniques. 
Finally, we discuss existing challenges and point out promising research directions of MLLMs.

\end{sloppypar}

\end{abstract}

\begin{document}

\begin{sloppypar}

\begin{CJK}{UTF8}{gbsn}
\section{Introduction}\label{sec0}
\begin{figure*}[ht]
\centering
\includegraphics[width=0.95\textwidth]{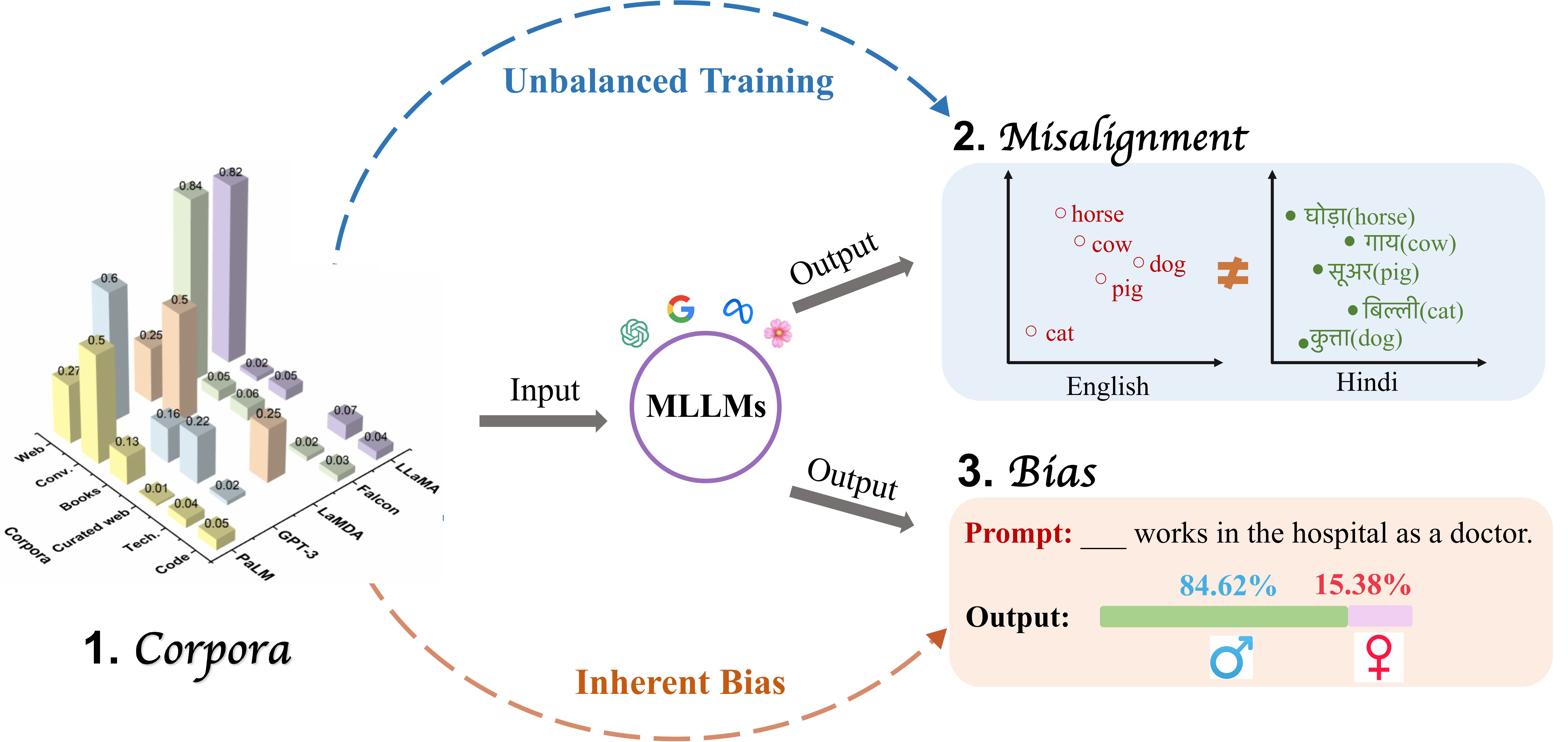}
\caption{An illustration of the relationship between corpora, misalignment, and bias. The misalignment and bias produced by MLLM arise in part from the bias and imbalanced language proportions of the training corpora.}
\label{fig-0}
\end{figure*}

The rapid development of Large Language Models (LLMs) has 
brought about a paradigm shift and revolution 
in the field of Natural Language Processing (NLP). 
This innovative approach trains a transformer-based model \cite{x1} on extensive volumes of data and then 
leverages fine-tuning or prompt learning to facilitate the model's adaption to a wide variety of tasks. 
Based on the foundation of LLMs, large-scale Multilingual MLLMs (MLLMs), 
such as mBERT \cite{x2}, XLM \cite{x3}, mT5 \cite{x4}, BLOOM \cite{x5} and LLaMA \cite{q5}, 
have been developed to tackle multilingual NLP tasks. 
MLLMs are pre-trained on a concatenation of texts in multiple languages 
with the hope that low-resource languages may benefit from high-resource languages 
due to linguistic similarities and shared representations inherent within language pairs.

Compared to LLMs, MLLMs require larger multilingual corpora that cover more languages 
to ensure applicability and fairness across different languages in downstream tasks. 
MLLMs are trained to understand and capture the structures and patterns of multiple languages. 
For instance, pre-trained on data from 104 languages, BLOOM supports 46 languages, 
covering the eight most widely spoken languages in the world \cite{x5}.
Numerous MLLMs have been proposed in the past 5 years,
which differ in the architecture (e.g., number of layers, parameters, etc), data used for pre-training (Wikipedia, Common Crawl, etc), 
and the number of languages involved (ranging from 12 to 110). 
However, it is uncertain how much cross-lingual transfer learning capability MLLMs have to support unseen languages or low-resource languages
during pre-training.
As a result, section \ref{sec-1} first starts by providing an overview of MLLMs, which contains key evolutions, techniques, and a detailed analysis of MLLMs’ multilingual capacities. 
Despite the success of MLLMs, existing MLLMs still face numerous issues and challenges, which can be summarized as three aspects: corpora, alignment, and bias. 
As shown in Figure \ref{fig-0}, 
training corpora of MLLMs heavily influence their capability. 
On the one hand, the unbalanced corpora and training lead to misalignment of MLLMs among different languages; On the other hand, inherent bias within corpora induces MLLMs to produce biased output. 
Therefore, 
this paper focuses discussion around the three aspects of corpora, alignment, and bias.

Firstly, 
MLLMs heavily rely on multilingual corpora to enhance their performance. 
For example, among the training corpus of ChatGPT, the English corpus accounts for 92.099\%, and the Chinese only accounts for 0.16\%, 
so its dialogues in the English context are much higher than those in other languages in terms of quality and speed. 
However, the size of available corpus resources for different languages varies greatly, 
and most of the existing annotated datasets are mainly focused on a few languages, limited the cross-lingual transfer effectiveness
from high-resource languages to languages that are unseen during training.
Furthermore, MLLMs suffer from what Conneau et al. \cite{x6} call
\textit{the curse of multilinguality}: 
more languages lead to better cross-lingual performance on low-resource languages up until a point, 
after which the overall performance of MLLMs on monolingual and cross-lingual benchmarks will decrease. 
In conclusion, the scale, quality, and diversity of corpora have a significant impact on the performance of MLLMs.
Therefore, section \ref{sec-2} presents a survey of the representative multilingual training corpora of MLLMs, 
offering insights into their language distribution, data source, and language coverage.

Secondly, MLLMs still struggle to learn a universal language representation for diverse languages.
The language misalignment issues exist.
Aligning the representation of diverse languages acts as  an integral part of NLP's multilingual tasks and applications\cite{m15}, and under-representation of low-resource languages leads to MLLMs' poor performance on these languages.
Inspired by the impressive performance of monolingual representation models like Word2vec \cite{x7} and GloVe \cite{x8},
recent research has made great progress in multilingual representation.
In section \ref{sec-3}, 
we review previous research on multilingual representations and 
classify them into three categories:
static multilingual representation,
contextual multilingual representation, and
combined multilingual representation.
We also analyze the impact of various factors on multilingual alignment performance,
including initial alignment solution, 
linearity of mapping function, 
language typological distance,
and pre-training data and settings of MLLMs.

Thirdly, MLLMs are prone to produce harmful outcomes and social bias 
\cite{mo21} in part due to bias is naturally present in cross-cultural datasets
and the design of MLLMs' modeling processes \cite{x10}.
Previous studies have explored bias in various NLP tasks and 
demographic groups，but are largely specific to English-based models \cite{x11,x12},
which cannot be generalized to other languages.
What are the types of bias in existing MLLMs? 
What are the main de-biasing techniques available for MLLMs? 
Does the removal of these biases affect LLMs’ performance? 
What are the existing bias evaluation datasets for MLLMs? 
These are very worthwhile research questions.
This survey tries to answer these questions and offers valuable insights for bias on MLLMs.

The contributions of this survey are as follows:
\begin{itemize}
    \item We present an overview of MLLMs and analyze the 
    language imbalance challenge within MLLMs, 
    their capacity to support low-resource languages
    and their potential for cross-lingual transfer learning.
     \item We provide an overview of the multilingual datasets and corpora utilized by existing MLLMs, offering
    a comprehensive insight into the language distribution within these training corpora.
    \item We survey the existing studies on multilingual representations 
    and explore whether the current MLLMs can learn a universal language representation.
    \item Our survey delves into bias within MLLMs,
    seeking to address essential questions such as 
    identifying the types of bias present in current MLLMs,
    exploring prominent de-biasing techniques,
    and summarizing available bias evaluation datasets for MLLMs.

\end{itemize}

\section{Overview of MLLMs} \label{sec-1}
\begin{figure*}[ht]
\centering
\includegraphics[width= 0.95\textwidth]{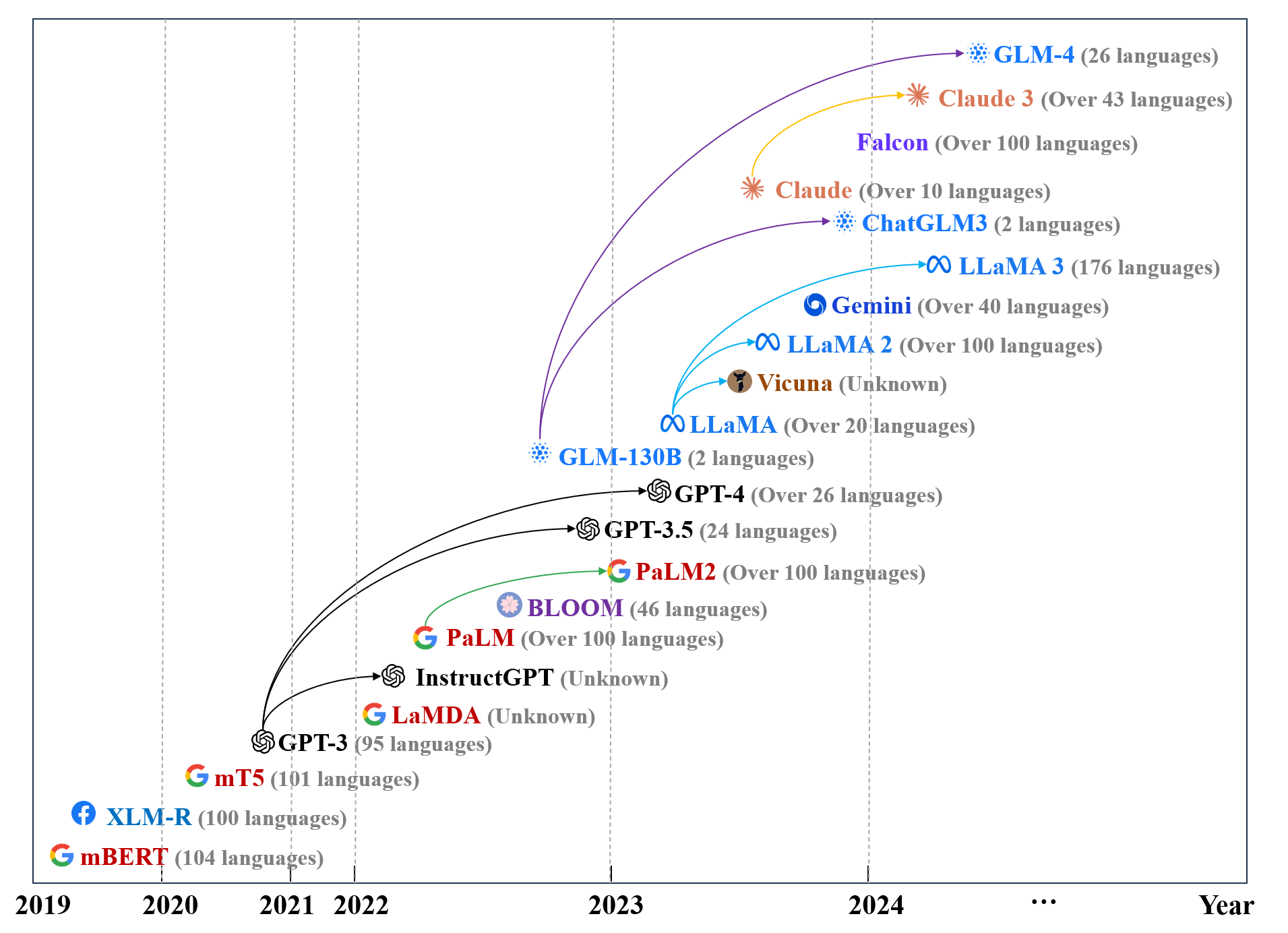}
\caption{An illustration of the evolution roadmap of current multilingual LLMs,
presenting their release year,
the number of supported languages and release relationship.
`Unknown' indicates the model has not 
disclosed the language proportion in its training data.}
\label{fig-llmroadmap}
\end{figure*}

This section provides a brief overview of MLLMs, tracing their evolution from monolingual LLMs to multilingual LLMs,
as dispicted in Figure \ref{fig-llmroadmap}.
It then illustrates the key techniques that contribute most to the success of MLLMs, as well as the multilingual capacities of MLLMs.

\subsection{Evolution of MLLMs}\label{1-1}

\subsubsection{Monolingual Evolution}\label{1-1-1}
The development of monolingual LLMs has made great progress in understanding and generating human languages. These models employ a transformer-based architecture that is first pre-trained on a large corpus of texts, followed by fine-tuning or prompt learning to enhance models’ performance on specific tasks or languages.

The representative monolingual LLMs are the BERT series and GPT series. With the success of BERT \cite{x2}, BERT-variants models have been developed for specific languages, such as FlauBERT for French \cite{a14}, BERTje for Dutch \cite{a16}, AraBERT for Arabic \cite{a20}. The GPT series, evolving from GPT-1 to GPT-2, GPT-3, and beyond \cite{m24, x13, x14, q2, mo4}, has experienced growth in parameters and training corpora, e.g., parameters ranging from hundreds of millions (GPT-1) \cite{m24} to 1.5 trillion (GPT-3) \cite{x14}. This progression empowers the models with improved sophisticated language understanding and generation capabilities. T5 model \cite{a39} introduces a unified framework that converts a variety of tasks into a text-to-text format by prepending a unique prefix to the input for each task. BART \cite{a64} is a sequence-to-sequence model, not an auto-regressive model like GPT-2 or an auto-encoder model like BERT, and thus it is particularly effective on comprehension tasks like summarization.

Monolingual LLMs have seen advancements in transformer-based architecture and pre-training strategies but are language-dependent. Training such language-specific LLMs is only feasible for a few languages with necessary corpora.

\subsubsection{Multilingual Evolution}\label{1-1-2}

Multilingual LLMs build upon the foundation of monolingual LLMs to learn universal language patterns from extensive unlabeled data across multiple languages. MLLMs have advanced to overcome linguistic limitations and enhance low-resource languages by leveraging shared vocabulary and genetic relatedness from high-resource languages \cite{a69}.

Following the success of monolingual BERT, multilingual BERT (mBERT) \cite{x2} was the first released MLLM by following the training procedure of BERT but on multilingual Wikipedia text corpora including 104 languages. Other MLLMs such as XLM-R \cite{x6}, mBART\cite{a35} and mT5 \cite{x4} follow the step of mBERT, further exploring the capacities and limitations of MLLMs across languages. Studies reveal that MLLMs work surprisingly well on cross-lingual tasks even without direct cross-lingual supervision like parallel or comparable data \cite{a67, a68}.

The further development trend has led to a tremendous increase in model parameters and data scale, resulting in enhancements that promote multilingual capabilities. For example, PaLM with 540 billion (540B) parameters yields impressive capabilities on the multilingual benchmarks by training on a mixture of multilingual versions of Wikipedia and dialogue data, including 124 languages \cite{q4}. 
With the early successful attempts, the autoregressive language modeling and prompt learning paradigm represented by the GPT series has received much attention and follow-up from major companies and universities. Thus, more MLLMs (e.g., InstructGPT \cite{q2}, LaMDA \cite{q1}, OPT \cite{a70}, BLOOM \cite{x5}, LLaMA \cite{q5}) have been proposed to achieve breakthrough performance in a range of multi-step reasoning tasks over multiple languages. 
In addition to the GPT series and its derivative models, 
numerous other models have been proposed to boost the development of LLMs. Examples include GLM \cite{a66, a37}, Vicuna \cite{q6}, Gemini \cite{mo18}, and several others. 


The development of MLLMs has been guided by several tendencies: 
(1) parameter growth; (2) linguistic diversity; (3) multimodal unification.
Regarding (1),
MLLMs have expanded to hundreds of billions of parameters or even trillions.
Increasing the size of parameters brings clear benefits,
such as alleviating hallucination phenomena heavily present in minor-parameter (e.g., 7B, 13B) models.
However, there's a limit to the amount of text data available online, and obtaining high-quality data is becoming increasingly challenging, which might slow down the parameter growth of MLLMs.
Regarding (2),
most high-resource languages belong to similar language families, thus sharing numerous linguistic features. 
Disregarding this diversity inevitably leads to poor generalizability and language-specific biases \cite{a71}.
Recent work in MLLMs has focused on addressing this issue 
as low-resource and unseen languages still account for a large proportion of the world's languages.
Regarding (3), 
multimodal MLLMs are a growing focus of research, 
realizing a variety of specific real-world needs by unifying diverse types of modalities (i.e., text, image, and speech). 
Additionally, current research aims to extend MLLMs to accommodate more modalities like web pages, heat maps, graphs, and tables, 
thereby increasing the model's generality and applicability \cite{a72}.

Table~\ref{tab1} summarizes the representative MLLMs in recent years, divided into two categories: monolingual and multilingual, showing the evolution of MLLMs from multiple perspectives in chronological order of release.


\begin{table*}[htbp]
  \centering
  \setlength{\abovecaptionskip}{5pt}
  \caption{An overview of representative MLLMs in recent years, including their release time, publishing authority, maximum available parameters, maximum supported context length, pre-training file size, architecture, base model, pre-training function, publicly available, and modal.}

    \scalebox{0.75}{
    \renewcommand\arraystretch{0.7}
    \begin{tabular}{lcccccccccc}

    \toprule
    \textbf{Model} & 
    \makecell{\textbf{Release} \\ \textbf{Time}} & 
    \makecell{\textbf{Publishing} \\ \textbf{Authority}} & \textbf{Params} & \makecell{\textbf{Context} \\ \textbf{Length}} & 
    \makecell{\textbf{Pre-training} \\ \textbf{File Size}} & 
    \textbf{Architecture} &
    \makecell{\textbf{Base} \\ \textbf{Model}} &
    \makecell{\textbf{Pre-training} \\ \textbf{Function}} &
    \makecell{\textbf{Publicly} \\ \textbf{Available}} & 
    \textbf{Modal} \\
    \midrule
    \multicolumn{11}{c}{{Monolingual}} \\
    \midrule
    GPT-1 \cite{m24}& Jun-18 & OpenAI & 117M  & 2K    & -      & Decoder-only & GPT   & LM    & Open  & Text \\
    \midrule
    BERT \cite{x2} & Oct-18 & Google & 340M  & 2K    & 1.3GB & Encoder-only & -      & \makecell{Seq2Seq \\ MLM} & Open  & Text \\
    \midrule
    GPT-2 \cite{x13}& Feb-19 & OpenAI & 1.5B  & 2K    & 40GB  & Decoder-only & GPT   & LM    & Open  & Text \\
    \midrule
    T5 \cite{a39}   & Oct-19 & Google & 11B   & 2K    & 21GB  & En-decoder & -      & \makecell{Seq2Seq \\ MLM}
    & Open  & Text \\
    \midrule
    GPT-3 \cite{x14} & May-20 & OpenAI & 175B  & 2K    & 570GB & Decoder-only & GPT   & LM    & Closed & Text \\
    \midrule
    Gopher \cite{q25} & Dec-21 & DeepMind & 280B  & 2K    & -      & Decoder-only & -      & LM    & Open  & Text \\
    \midrule  
    \multicolumn{11}{c}{{Multilingual}} \\
    \midrule
    mBERT \cite{x2} & Jul-19 & Google & 172M  & 2K    & -      & Encoder-only & BERT  & MLM   & Open  & Text \\
    \midrule
    XLM-R \cite{x6} & Nov-19 & Facebook & 550M  & 2K    & -      & Encoder-only & -      & TLM   & Open  & Text \\
    \midrule
    mBART \cite{a35} & Jan-20 & Facebook & 680M  & 2K    & -      & En-decoder & BART  & DAE   & Open  & Text \\
    \midrule
    mT5 \cite{x4}  & Oct-20 & Google & 13B   & 2K    & -      & En-decoder & T5    & \makecell{Seq2Seq \\ MLM} 
    & Open  & Text \\
    \midrule
    LaMDA  \cite{q1} & Jan-22 & Google & 137B  & 32K   & - & Decoder-only & -      & LM    & Open  & Text \\
    \midrule
    PaLM  \cite{q4}& Apr-22 & Google & 540B  & 2K    & -      & Decoder-only & -      & LM    & Closed & Text \\
    \midrule
    BLOOM  \cite{x5}& Jul-22 & BigScience & 176B  & 2K    & 350GB & Decoder-only & -      & LM    & Open  & Text \\
    \midrule
    GLM-130B \cite{a37}& Aug-22 & ZHIPU & 130B  & 2K    & -      & En-decoder & GLM   & ABI   & Closed & Text \\
    \midrule
    FLAN-T5 \cite{mo9}& Oct-22 & Google & 11B   & 2K    & 17.3MB & En-decoder & T5    & LM    & Open  & Text \\
    \midrule
    GPT-3.5 \cite{x14} & Nov-22 & OpenAI & 175B  & 2K    & -      & Decoder-only & GPT   & LM    & Closed & Text \\
    \midrule
    ChatGPT \cite{mo3} & Nov-22 & OpenAI & 175B  & 2K    & -     & Decoder-only & GPT-3.5 & LM    & Open  & Text \\
    \midrule
    LLaMA \cite{q5} & Feb-23 & Meta  & 65B   & 4K    & 120GB & Decoder-only & -      & LM    & Open  & Text \\
    \midrule
    ChatGLM \cite{q25} & Mar-23 & ZHIPU & 130B  & 2K    & 8GB   & En-decoder & GLM   & ABI   & Open  & Text \\
    \midrule
    PaLM-E \cite{mo10}& Mar-23 & Google & 562B  & 2K    & -      & Decoder-only & PaLM  & LM    & Open  & Text, Image \\
    \midrule
    Alpaca \cite{mo11} & Mar-23 & StandFord & 7B    & 2K    & 14GB  & Decoder-only & LLaMA & LM    & Open  & Text \\
    \midrule
    GPT-4 \cite{mo4} & Mar-23 & OpenAI & -      & 8K    & -      & Decoder-only & GPT   & LM    & Closed & Text, Image \\
    \midrule
    PanGu-Σ \cite{mo12} & Mar-23 & Huawei & 1085B & -     & -     & Decoder-only & PanGu-α & LM    & Closed & Text \\
    \midrule
    Pythia \cite{mo13}& Apr-23 & EleutherAI & 12B   & 2K    & 24GB  & Decoder-only & -     & LM    & Open  & Text \\
    \midrule
    PaLM 2 \cite{q7} & May-23 & Google & 340B  & 2K    & -      & Decoder-only & PaLM  & LM    & Closed & Text \\
    \midrule
    ChatGLM2 \cite{a66} & Jun-23 & ZHIPU & 12B   & 4K    & -      & En-decoder & GLM   & ABI   & Closed & Text \\
    \midrule
    Vicuna \cite{q6}& Jun-23 & LMSYS & 33B   & 2K    & 65GB & Decoder-only & LLaMA & LM    & Open  & Text \\
    \midrule
    LLaMA 2 \cite{q103} & Jul-23 & Meta  & 70B   & 4K    & 129GB & Decoder-only & LLaMA & LM    & Open  & Text \\
    \midrule
    Bard  \cite{mo20} & Jul-23 & Google & -     & -     & -      & Decoder-only & LaMDA & LM    & Open  & Text, Image \\
    \midrule
    Baichuan \cite{mo14} & Jul-23 & BAICHUAN & 13B   & 4K    & 26.6GB & Decoder-only & -      & LM    & Open  & Text \\
    \midrule
    GPT-4V \cite{mo4}& Sep-23 & OpenAI & -      & 32K   & -      & Decoder-only & GPT-4 & LM    & Closed & Text, Image \\
    \midrule
    ChatGLM3 \cite{a37} & Oct-23 & ZHIPU & 6B    & 32K   & 12GB  & En-decoder & GLM   & ABI   & Open  & Text \\
    \midrule
    GPT-4 Turbo \cite{mo4} & Nov-23 & OpenAI & -     & 128K  & -      & Decoder-only & GPT-4 & LM    & Closed & \makecell{Text, Image, \\Speech} \\
    \midrule
    Gemini-ultra \cite{mo18} & Dec-23 & DeepMind & 300B  & 32K   & -      & Decoder-only & -      & LM    & Closed & Text, Image \\
    \midrule
    Phi-2 \cite{mo19}& Dec-23 & Microsoft & 2.7B  & 2K    & 5.4GB & Decoder-only & -      & LM    & Open  & Text \\
    \midrule
    GLM-4 \cite{mo16} & Jan-24 & ZHIPU & -      & 128K  & -      & En-decoder & GLM   & ABI   &  Closed & Text, Image \\
    \midrule
    Claude 3 \cite{k2} & Mar-24 & Anthropic & -   &  200K &  -   & Decoder-only & Claude  &  LM  &  Closed  & Text, Image   \\
    \midrule
    LLaMA 3 \cite{k1} & Apr-24 & Meta &  70B  & 8K  &    140GB   &  Decoder-only &  LLaMA  &  LM  & Open & Text  \\
    \bottomrule
        \end{tabular}%

    }
  \label{tab1}%
\end{table*}%

\subsection{Key Techniques of MLLMs}\label{1-2}

Transformer architecture, pre-training technique, and reinforcement learning with human feedback 
are the key techniques for MLLMs. 
In this section, we will present the key ideas behind these techniques.

\subsubsection{Transformer Architecture}\label{1-2-1}

Transformer architecture, first introduced in 2017, has become the foundation of MLLMs owing to its suitability for parallel computing and flexibility for diverse model design.
Transformer architecture consists of two main modules, Encoder and Decoder, along with a self-attention mechanism within each module.
The encoder, using stacked multi-head self-attention layers, encodes the input sequence and generates latent representations. 
In contrast, the decoder employs cross-attention to utilize the encoder’s latent representations, attending to them while autoregressively generating the target sequence \cite{a62}.

MLLMs can be categorized into three groups based on the underlying transformer structure: 
\begin{itemize}
\item{\textbf{Encoder-only} (e.g., BERT )}: MLLMs with encoder-only architecture can effectively handle long-range dependencies within the input sequences, making them well-suited for the analysis and classification of textual content, including tasks like sentiment analysis and named entity recognition.

\item{\textbf{Decoder-only} (e.g., GPT)}: MLLMs with decoder-only architecture are mainly designed to generate sequences of language texts. They predict the next token based on contextual information from the current and preceding steps.

\item{\textbf{Encoder-decoder hybrid} (e.g., GLM)}: MLLMs with encoder-decoder architecture enable themselves to process sequential data and generate accurate and coherent outputs that excel in tasks such as text generation, and summarization.

\end{itemize}

\subsubsection{Pre-training Technique}\label{1-2-2}
Pre-training technique aims to learn universal language representations from billion-scale unlabeled corpora (e.g., Wikipedia, webpages, news, etc.) and 
then initializes the parameters of the Transformer-based MLLMs.
This approach reduces the reliance on massive parallel corpora, helping MLLMs generate similar representations in a common vector space for similar sentences and words 
(or words in similar context) across languages\cite{a13}.

The benefits of pre-training technique can be attributed to two key factors: Paradigm and Task. 
Pre-training paradigms have been proposed to capture linguistic patterns in the training data and adapt MLLMs to downstream tasks, including ``pre-training + fine-tuning" and ``pre-training + prompting". The former representative models are BERT \cite{x2}, GPT-2 \cite{x13},  while the latter presentative models like GPT-3 \cite{x14}. 
Pre-training tasks improve the ability of MLLMs to encode and generate coherent multilingual text. 

When learning the universal representation of language, 
pre-training tasks play a crucial role
and the widely used pre-training tasks include probabilistic language modeling (LM),
masked language modeling (MLM), next sentence prediction (NSP), and Denoising autoencoder (DAE).
Probabilistic LM is a fundamental task in NLP, 
estimating the probability distribution of sequences of words in a language.
In practice, LM typically involves auto-regressive LM or unidirectional LM. 
MLM has emerged as a novel pre-training task to overcome the drawback of the standard unidirectional LM. 
By masking certain tokens in a sequence and predicting them based on context, 
MLM encourages models to learn bidirectional representations, capturing dependencies from both left and right contexts.
Punctuations are the natural separators of text data. So, it is reasonable to construct pre-training methods by utilizing them. 
NSP is just a great example of this. 
NSP encourages the model to understand the contextual coherence and relationships between sentences. 
DAE takes a partially corrupted input and aims to recover the original undistorted input. 
Specific to language, a sequence-to-sequence model, such as the standard Transformer, is used to reconstruct the original text. 
Eq.\ref{eq:1} summarize loss function $\mathcal{L}$ of these pre-training tasks, where $\text{x}=[{{x}_{1}},{{x}_{2}},\ldots ,{{x}_{T}}]$ denotes a sequence of tokens \cite{a61}.
\begin{equation}\label{eq:1}
\begin{aligned}
&{{\mathcal{L}}_{\text{LM}}}=-\sum\limits_{t=1}^{T}{\log p({{x}_{t}}|{{\text{x}}_{<t}})} \\
&{{\mathcal{L}}_{\text{MLM}}}=-\sum\limits_{\hat{x}\in m(\text{x})}{\log p(\hat{x}|{{\text{x}}_{\backslash m(\text{x})}})} \\ 
&{{\mathcal{L}}_{\text{NSP}}}=-\log p(t|\text{x,y}) \\
&{{\mathcal{L}}_{\text{DAE}}}=-\sum\limits_{t=1}^{T}{\log p({{x}_{t}}|\hat{\text{x}},{{\text{x}}_{<t}})} \\
\end{aligned}
\end{equation}
where the symbols and their definitions are listed in Table \ref{Notation}.

\vspace{-0.2cm}
\begin{center}
\renewcommand{\arraystretch}{0.8}
    \begin{table}[!ht]
        \caption{Summary of key notations.}
        \label{Notation} 
        \centering

        \begin{tabular}{p{2cm}<{\centering}p{5.5cm}}
        \toprule 
         \textbf{Symbol}& \textbf{Description}   \\[5pt]
        \midrule
        \text{x}& $\text{x}=[{{x}_{1}},{{x}_{2}},\ldots ,{{x}_{T}}]$ \\ [5pt]
        \midrule
        m(\text{x})& The masked tokens \\ [5pt]
        \midrule
        ${\text{x}}_{\backslash m(\text{x})}$& The rest tokens  \\ [5pt]
        \midrule
        $\hat{x}$& Random perturbation text \\ [5pt]
        \midrule
        \text{y}& Target label \\ [5pt]
        \midrule
        $t$& If \text{x} and \text{y} are continuous segments from corpus, $t = 1 $ \\ [5pt]
        \midrule
        $\mathcal{P}$& The prompt \\ [5pt]
        \midrule
        ${{y}_{w}}$& The better model response \\ [5pt]
        \midrule
        ${{y}_{l}}$& The worse model response \\ [5pt]
        \midrule
        ${r}_{\theta }$& The output of reward model \\ [5pt]
        \midrule
        $\theta$& Weight in policy optimization network \\ [5pt]
        \midrule
        $J(\theta)$& The objective function \\ [5pt]
        \midrule
       $\nabla_\theta J(\theta)$& The gradient of objective function \\ [5pt]
    \bottomrule
        \end{tabular}   
    \end{table}
\end{center}
\vspace{-1.7cm}

\subsubsection{Reinforcement Learning with Human Feedback}\label{1-2-3}

MLLMs may generate inaccurate or harmful outputs due to their probabilistic statistical text generation mechanism \cite{a50}.
Reinforcement Learning from Human Feedback (RLHF) and its variants \cite{a53, a54, a55, a56, a57} 
have been proposed to fix this by optimizing MLLMs with human feedback, 
aligning them better with human values in three fundamental dimensions: helpfulness, honesty, and harmlessness \cite{a51}.

Essentially, RLHF has three core steps \cite{a59,a60}:
\begin{enumerate}
\item{\textbf{Pre-training a Language Model}}: 
To pre-train MLLMs, extensive prompts, and multilingual datasets are utilized as examples,
teaching the model how to respond appropriately in a specific context.
\item{\textbf{Training a Reward Model}}: Prompts serve as input to MLLMs, 
where $K$ pairs of \{prompt, response\} are manually scored by 
human evaluators to align with human preferences. 
The rankings of \{sample, reward\} pairs are normalized 
into a scalar reward signal for training the Reward Model (RM). 
The loss function $\mathcal{L}_{\theta}$ is defined as follows, 
where $\mathcal{P}$ is the prompt, ${{y}_{w}}$ and ${{y}_{l}}$ denote 
the better and worse model responses respectively, and ${{r}_{\theta }}(\mathcal{P},y)$ is the output of the RM.
{\small
\begin{equation}
    \mathcal{L}_{\theta}=-\frac{1}{(_{2}^{K})}{{E}_{(\mathcal{P},{{y}_{w}},{{y}_{l}})\sim D}}[\log (\sigma ({{r}_{\theta }}(\mathcal{P},{{y}_{w}})-{{r}_{\theta }}(\mathcal{P},{{y}_{l}})))]
\end{equation}
} where the symbols and their definitions are listed in Table \ref{Notation}.

\item{\textbf{Fine-tuning with Reinforcement Learning}}: 
MLLMs are optimized using Proximal Policy Optimization (PPO) \cite{a52} or 
a similar reinforcement learning (RL) algorithm such as A2C \cite{a58}.
These RL algorithms incorporate human-generated reward signals to 
apply gradient strategies for learning human feedback.
\end{enumerate}

\begin{figure*}[!t]
\centering
\includegraphics[width=0.9\textwidth]{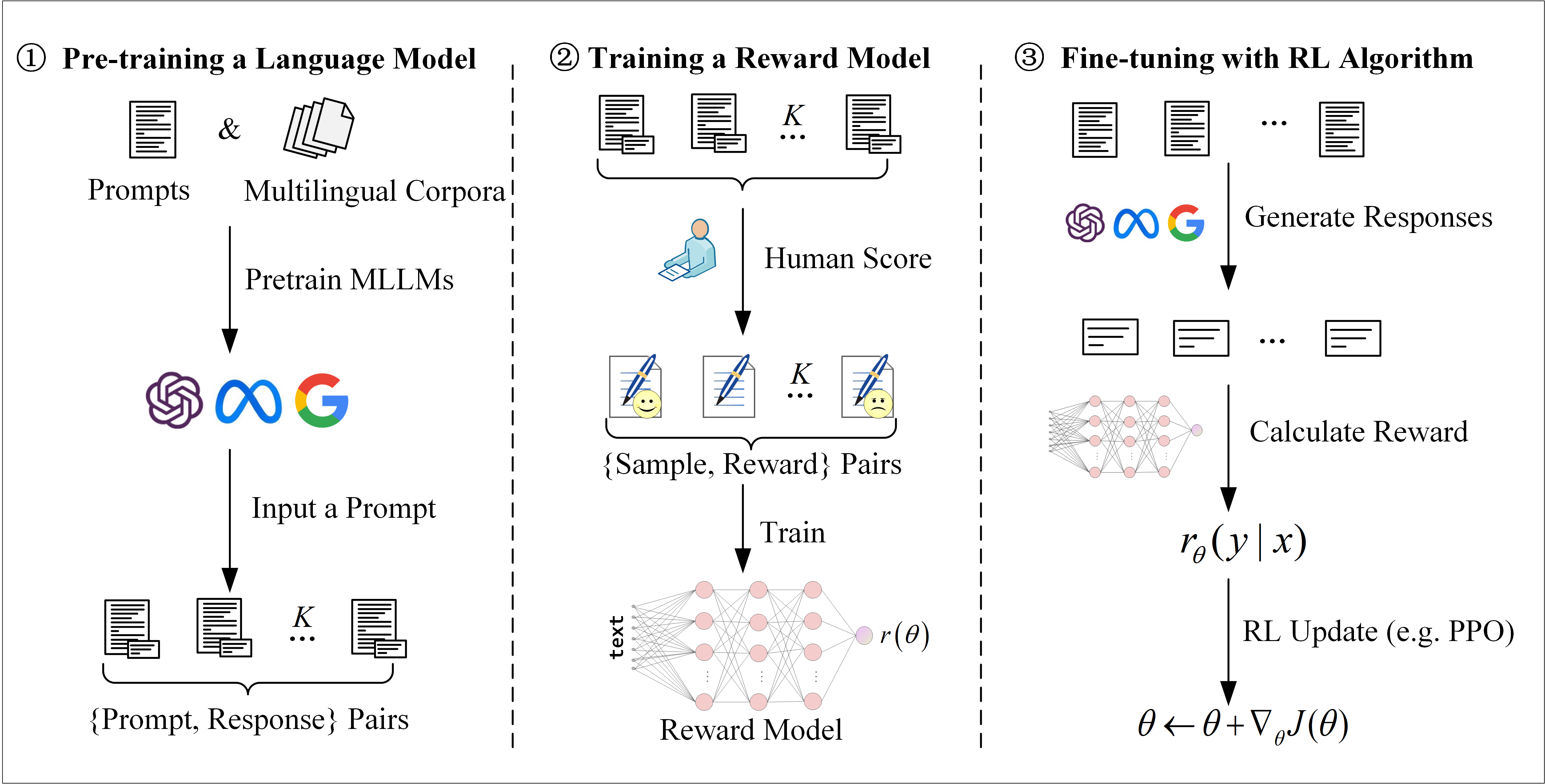}
\caption{Diagram illustrating the RLHF procedure, which consists of three key steps: (1) Pre-training a LM using the labeled prompt-response dataset, (2) Training a Reward Model based on scores provided by human evaluators for LM's generation, and (3) Fine-tuning with a Reinforcement Learning (RL) algorithm, which helps to update parameters in the LM based on the feedback from RM.}
\label{fig-1-2}
\end{figure*}

\subsection{Multilingual Capacities of MLLMs}\label{1-3}
Pre-training MLLMs on extensive multilingual data enhances their 
multilingual capacities
and cross-lingual transfer learning (CLTL) from one language to another.
However, MLLMs still face challenges in training with multilingual corpora and their exact
CLTL capabilities remain largely unexplored. 
This section focuses on these two concerns.

\subsubsection{Challenges brought by Multilingual Corpora}\label{1-3-1}
Three challenges arise from multilingual corpora training. 
Firstly, 
while MLLMs outperform monolingual LLMs
in downstream tasks for high-resource languages, 
their performance on low-resource languages remains unsatisfactory due to limited annotated data.
Secondly,
the ``curse of multilinguality" phenomenon in MLLMs worsens this situation.
Supporting more languages can lead to a significant decline in performance for 
low-resource languages, making them victims of this curse \cite{a47}.
Thirdly,
the distribution of languages in the pre-training corpora is highly skewed towards English,
further complicating efforts to address the ``curse of multilinguality" phenomenon.

To mitigate these challenges, 
two approaches have been proposed.
One involves fine-tuning existing MLLMs to suit the linguistic features of 
low-resource languages \cite{a6}. 
However, this method is constrained by 
the demand for extensive specific-task annotated training data \cite{a7}. 
Alternatively, another approach
is to pre-train monolingual LLMs specifically for low-resource languages \cite{a8}.
This method allows models to learn from diverse sources and contexts within the target language
without requiring costly annotated data. 
Therefore, MLLMs trained by this approach exhibit superior performance on low-resource languages
compared to the aforementioned fine-tuning approach.
For example, Torge et al. \cite{a12} pre-trained monolingual RoBERTa models for Czech and Polish, 
as well as a bilingual model for Czech-Polish,
which demonstrated superior performance to the current state-of-the-art multilingual model, XLM-R, across various 
downstream tasks.
Recently, there has been a growing interest in developing
low-resource language models to meet the demands of morphologically rich, low-resource languages.
Examples include language-specific BERT models like FlauBERT for French \cite{a14}, BERTje for Dutch \cite{a16}, 
and FinBERT for Finnish \cite{a17}, 
among others \cite{a13}.

The main reason for MLLMs' poor performance on low-resource languages is 
the skewed distribution of languages in the pre-training data. 
Therefore, techniques have been proposed to address this issue.
Data sampling techniques like exponential weighted smoothing \cite{x6}
help prevent the under-representation of low-resource languages,
while vocabulary augmentation approaches \cite{a49} enrich the model's vocabulary by 
inducing new tokens of unseen languages during training.
Moreover, research has also attempted to tackle the language imbalance.
Choenni et al. discovered that languages influence each other during the pre-training 
phase
and MLLMs benefit from reinforcement or complementary learning \cite{a43}.
Wang et al. emphasized the significance of imbalanced learning algorithms 
in Vision-Language models (VLMs) \cite{a44}.
For example, CLIP model demonstrated an improvement from 5\% to 69\% on iNaturalist dataset 
by adopting imbalanced methods. 
Jiang et al. proposed a data augmentation pipeline to address imbalance in social media data,  
effectively handling multiclass problems\cite{a45}. 

\subsubsection{Cross-lingual Transfer Learning brought by Multilingual Corpora}\label{1-3-2}

MLLMs can facilitate CLTL from 
one language to another.
This naturally raises the question of 
how much CLTL capability that MLLMs possess to support these unseen languages or low-resource languages
during pre-training.


Research has been dedicated to exploring the cross-lingual transferability of MLLMs through zero-shot learning.
Lin et al. \cite{mo2} trained 4 multilingual generative language models and 
examined their zero-shot and in-context few-shot learning capabilities in a wide range of tasks.
They found that these models can achieve cross-lingual few-shot learning in non-English languages
without requiring source-to-target language translation. 
Tian et al. \cite{a23} found that MLLMs exhibit strong rumour detection performance in zero-shot cross-lingual transfer learning. 
What's more, MLLMs showed surprisingly strong multilingual reasoning abilities even in under-represented languages such as Bengali and Swahili \cite{a42}.

To further improve the transfer learning performance of MLLMs on unseen or low-resource languages, 
as these languages still account for a significant portion of the world's languages,
MLLMs are pre-trained to learn languages from the same linguistic family or branch \cite{a5,a11}. 
MLLMs trained on a small amount of data from 
genetically related languages 
could achieve performance comparable to the ones trained on large but unrelated data \cite{a5}.
MLLMs trained on only low-resource languages
with small datasets, which are similar to each other, sometimes achieved better performance than
models trained on large datasets with high-resource languages \cite{a11}.
For example, the AfriBERTa model \cite{a11}, pre-trained on less than 1 GB of text data from 11 African languages, 
most of which belong to the Bantu branch of the Niger-Congo language family, 
demonstrated the effectiveness of scratching solely on low-resource languages without any high-resource transfer learning.

A prominent future concern will be how to improve the CLTL capacities of MLLMs.
Pikuliak et al. conducted a survey on existing cross-lingual transfer paradigms of MLLMs \cite{a48},
while Philippy et al. investigated various factors that impacted cross-lingual
transfer performance, including linguistic similarity, lexical overlap, model architecture, 
pre-training setting, and pre-training corpus size \cite{a22}.
Specifically, 
this avenue of research seeks to
investigate how and why MLLMs possess different CLTL abilities on various languages.
This pursuit holds the potential to leverage
CLTL capacities to mitigate the dependence on annotated data 
and maintain or even enhance the performance of MLLMs in well-trained or unseen languages.

\section{Multilingual Corpora and Datasets}\label{sec-2}
In this section, we delve into the widely utilized multilingual corpora that are associated with the training of MLLMs, 
and multilingual datasets oriented for downstream tasks.

Table  \ref{tab-2-1} 
summarizes the multilingual corpora that representative MLLMs trained on, offering insights into their language distribution, data source,  and language coverage. 

\begin{center}
    \begin{table*}
    \captionsetup{width=0.90\textwidth} 
        \caption{An overview of representative multilingual training corpora of MLLMs in recent years, including their corresponding model, language, language proportion, and source.}
        \label{tab-2-1} 
    \scalebox{0.75}{  
        \centering
    \renewcommand\arraystretch{0.65}

        \begin{tabular}{llll}
        \toprule 
        \textbf{Model} & \makecell[l]{\textbf{Language}}& \makecell[l]{\textbf{Language proportion}} & \makecell[l]{\textbf{Source}}  \\[10pt]
        \midrule
        \makecell[l]{mBERT \cite{x2}} &\makecell[l]{104 languages}& \makecell[l]{Unknown} &\makecell[l]{Wikipedia}\\ [10pt]
        \midrule
        \makecell[l]{XLM-R \cite{x6}} &\makecell[l]{100 languages}& \makecell[l]{English (12.56\%); Russian (11.61\%);\\Indonesian (6.19\%); Vietnamese (5.73\%);\\ Others (63.89\%)} &\makecell[l]{Generated using the open source; \\CC-Net repository}\\
        \midrule
        \makecell[l]{mT5 \cite{x4}} &\makecell[l]{101 languages}& \makecell[l]{English (5.67\%); Russian (3.71\%); \\Spanish (3.09 \%); German (3.05\%); \\Others (84.48\%)} &\makecell[l]{Common Crawl}\\
        \midrule
        \makecell[l]{GPT-3 \cite{x14}} &\makecell[l]{95 languages}& \makecell[l]{ English (92.7\%);  French (1.8\%);\\ German (1.5\%); Others (5.9\%)} &\makecell[l]{Common Crawl;  Wikipedia;\\Books1; Books2; WebText2}\\
        \midrule
        \makecell[l]{Gopher \cite{q25}} &\makecell[l]{51 languages}& \makecell[l]{Over 99\% English } &\makecell[l]{MassiveWeb (48\%); C4 (10\%); News (10\%);\\Books (27\%); GitHub (3\%); Wikipedia (2\%)}\\
        \midrule        
        \makecell[l]{LaMDA \cite{q1}} & \makecell[l]{Unknown}& \makecell[l]{Over 90\% English} &\makecell[l]{Public dialog data and other public\\web documents}\\
        \midrule
        \makecell[l]{PaLM \cite{q4}} & \makecell[l]{Over 100 languages}& \makecell[l]{English (77.98\%); German (3.50\%); \\French (3.25\%); Spanish (2.11\%); \\Others (13.15\%)} & \makecell[l]{Social media conversations (50\%);\\Filtered webpages (27\%); \\Books (13\%);  GitHub (5\%); \\Wikipedia (4\%); News (1\%)}\\
        \midrule
        \makecell[l]{BLOOM \cite{x5}} & \makecell[l]{46 languages}& \makecell[l]{English (30.03\%); Simplified Chinese (16.16\%);\\ French (12.9\%); Spanish (10.85\%); \\Portuguese (4.91\%); Arabic (4.6\%); \\Others (20.55\%)} & \makecell[l]{Web Crawl(38\%);\\BigScience Catalogue Data(62\%)}\\
        \midrule
        \makecell[l]{LLaMA \cite{q5}}  & \makecell[l]{Over 20 languages}& \makecell[l]{Over 67\% English} & \makecell[l]{Common Crawl (67.0\%); C4 (15.0\%); \\Github (4.5\%);Wikipedia (4.5\%); \\Books (4.5\%); ArXiv (2.5\%); StackExchange (2.0\%)}\\
        \midrule
        \makecell[l]{Vicuna \cite{q6}}& \makecell[l]{Unknown} & \makecell[l]{Unknown} & \makecell[l]{User-shared conversations from\\ShareGPT.com}\\
        \midrule
         \makecell[l]{Falcon \cite{q108}}& \makecell[l]{Over 100 languages} & \makecell[l]{Excluding English:\\Russian (13.19\%); German (10.81\%);\\Spanish (9.45\%); Others (66.55\%)} & \makecell[l]{Common Crawl}\\
        \midrule
         \makecell[l]{PaLM 2 \cite{q7}}& \makecell[l]{Over 100 languages} & \makecell[l]{Excluding English:\\Spanish (11.51\%); Chinese (10.19\%);\\Russian (8.73\%); Others (69.57\%)} & \makecell[l]{Web documents; Books; \\ Code; Mathematics; \\Conversational data}\\
         \midrule
         \makecell[l]{LLaMA 2 \cite{q103}}& \makecell[l]{Over 100 languages} & \makecell[l]{English (89.70\%); Unknown (8.38\%); \\German (0.17\%); France (0.16\%); \\Others (1.59\%)} & \makecell[l]{Publicly available sources excludes\\Meta user data}\\
         \midrule
         \makecell[l]{GPT-4 \cite{mo4}}& \makecell[l]{Over 26 languages} & \makecell[l]{Unknown} & \makecell[l]{Common Crawl;  Wikipedia;\\Books1; Books2; WebText2}\\
         \midrule
         \makecell[l]{LLaMA 3 \cite{k1}}& \makecell[l]{176 languages} & \makecell[l]{Over 5\% non-English} & \makecell[l]{  Publicly available sources excludes\\Meta user data}\\
        \midrule
         \makecell[l]{GLM-4 \cite{mo16}}& \makecell[l]{26 languages} & \makecell[l]{Mostly English and Chinese} & \makecell[l]{Webpages;Wikipedia;Books;\\Code;Research papers  }\\
         \midrule
         \makecell[l]{Claude 3 \cite{k2}}& \makecell[l]{Over 43 languages} & \makecell[l]{Unknown} & \makecell[l]{ Publicly available information on the Internet;\\Non-public data from third partie;\\Data provided by data labeling services and paid contractors;\\Data they generate internally}\\
        \midrule
         \makecell[l]{YAYI 2 \cite{YAYI2}}& \makecell[l]{Over 10 languages } & \makecell[l]{Chinese (41.5\%); English (40.4\%); \\French (2.5\%); Spanish (2.2\%); \\Others (25\%)} & \makecell[l]{Web pages; Social media;\\Books; Newspapers; Academic Papers }\\
         \midrule
         \makecell[l]{FuxiTranyu \cite{sun2024fuxitranyu}}& \makecell[l]{43 languages} & \makecell[l]{English (28.8\%);  Chinese (10.4\%); \\German (8.1\%); French (7.3\%); \\Others (45.4\%)} & \makecell[l]{Web (82\%); Code (9\%);\\Paper (3\%); Book (3\%);\\Encyclopedia(2\%); Report(1\%)}\\

    \bottomrule
        \end{tabular}   
        }
    \end{table*}
\end{center}
\vspace{-1.55cm}

MLLMs have a more extensive language coverage in their training data compared to LLMs. 
A significant portion of these training data originates from multilingual repositories like Common Crawl, Wikipedia, and web documents, 
encompassing a broad range of languages. 
These multilingual repositories are crucial for enhancing the cross-lingual capability of MLLMs. In this section, we discuss training data’s language composition from both a general perspective and a language family perspective.

\subsection{Multilingual Corpora in MLLMs}
\label{sub51}
First, we analyze the linguistic composition of MLLMs’ training data, investigating the total number of languages and different language proportions within each training corpora. Analysis reveals that most MLLMs are trained on corpora where English is the predominant language. Notably, several MLLMs, including GPT-3 \cite{x14}, Gopher \cite{q25}, LaMDA \cite{q1} and InstructGPT \cite{q2}, are trained on corpora where English comprises over 90\%. The overwhelming English texts in corpora lead to MLLMs’ English-centric ability. To alleviate this issue, some MLLMs are trained on corpora with more balanced language distribution. For example, the training data of BLOOM \cite{x5} covers 46 languages, with English comprising less than half. YAYI 2 \cite{YAYI2} makes great efforts to balance its training data, achieving a nearly 1:1 ratio between English and Chinese training data. Compared to its base model PaLM \cite{q4}, PaLM 2 \cite{q7} includes a higher percentage of non-English data, further enhancing its multilingual capabilities. We present the percentages of its non-English languages in the web documents subset of its pre-training corpus in Table \ref{tab-2-1}, as the language distribution for English was not published.

\begin{figure*}[!ht]
\centering
\includegraphics[width=1\textwidth]{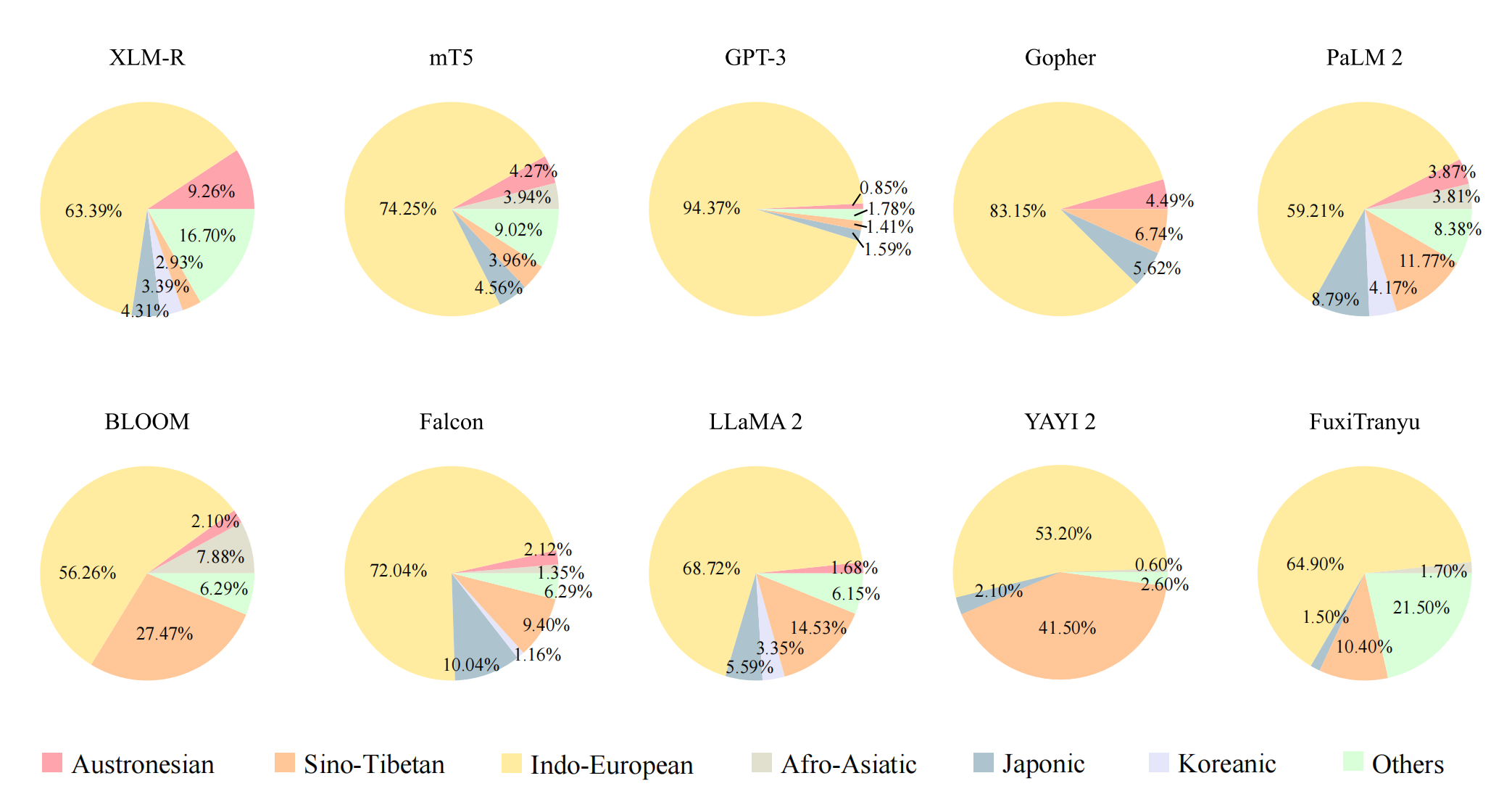}
\caption{This analysis excludes English and focuses on ratios of language families of languages (top 20) in MLLM's corpora. Note that Gopher only released the top 10 languages and FuxiTranyu only released the top 13 languages used in training corpora. What's more, some of the latest models like GPT-4 have not disclosed the proportion of their training data, so they aren't included in the chart.}
\label{fig-2}
\end{figure*}

Second, we explore the language composition of MLLMs’ training data from a language family perspective. Languages within the same language family share similar characteristics and MLLMs have better transfer performance on languages belonging to the same language family \cite{a13}. Thus, the proposition of language families in MLLMs’ training data can help us better understand the multilingual capabilities of MLLMs. What’s more, we can also leverage language families to observe the linguistic composition of the MLLMs’ training data. Since English is predominant in most MLLMs’ corpora, considering it in language family analysis would heavily favor the Indo-European language family to which English belongs. To gain a more detailed understanding of the language family proposition in MLLMs' corpora, we exclude English and focus on the top 20 prominent non-English languages of the training data and their corresponding language families. The distribution of language families of each MLLM is shown in Fig. \ref{fig-2}. 

Notably, French, German, Chinese, and Spanish emerge as the most prevalent languages in the training data. For example, French constitutes 1.8\% of the training corpora for GPT-3\cite{x14} and 12.9\% of the training corpora for PaLM\cite{q4}. French, German, and Spanish all belong to the Indo-European language family, which demonstrates that the Indo-European language family holds a prominent position in MLLMs' corpora, both in terms of quantity and linguistic diversity. An exception to this is Chinese, which belongs to the Sino-Tibetan language family while maintaining a significant presence in the training corpora. But in terms of linguistic diversity, the Sino-Tibetan language family in the training corpora, mainly consisting of the Chinese language, is much less diverse compared to the Indo-European language family. Besides Indo-European and Sino-Tibetan language families, some other language families are found in most MLLMs’ training corpora as well. Similar to Sino-Tibetan, they mainly contain only one language in training corpora. For example, Austronesian mainly includes Indonesian, Japonic mainly includes Japanese, and Koreanic mainly includes Korean. Apart from the lack of diversity within the same language family, there is also a lack of diversity across different language families in MLLMs’ training corpora. For example, despite Niger-Congo and Trans-New Guinea being among the largest language families in the world, they are notably absent from the top 20 languages in the training data. 

Through the above analysis of multilingual training corpora in MLLMs, we have derived the following key insights: MLLMs broaden language coverage beyond LLMs, yet English remains dominant in their training corpora. From a language family perspective, Indo-European languages occupy a prominent place in terms of both quantity and linguistic variety. Further work should consider a more comprehensive inclusion of language families and prioritize linguistic diversity within the same language family when training MLLMs.

\subsection{Multilingual Datasets for Downstream Tasks}
\label{sub52}

Multilingual datasets play a crucial role in fine-tuning MLLMs to be adaptive across various NLP tasks.
Table \ref{Representative Multilingual Datasets for Downstream Tasks} summarizes some representative multilingual datasets, including Multilingual Named Entity Recognition (Multilingual NER), Multilingual Sentiment Analysis (Multilingual SA), Cross-Lingual Information Retrieval (Cross-Lingual IR), and Multilingual Text Classification (Multilingual TC).

 \begin{center}
 \begin{table*}[h!]
 \setlength{\abovecaptionskip}{-5pt}
 \centering
 \captionsetup{width=1\textwidth} 
 \caption{An overview of representative multilingual datasets for downstream tasks in recent years, including their corresponding task, release name, language, size, and source.}
\scalebox{0.9}{

    \renewcommand\arraystretch{0.9}
 \begin{tabular}{p{2cm}p{3.3cm}p{3cm}p{3cm}p{4.5cm}}
 \label{Representative Multilingual Datasets for Downstream Tasks} \\
 \toprule
 \textbf{Task} &  \textbf{Dataset} & \textbf{Language}  & \textbf{Source} &  \textbf{Size}\\
 \hline
 Multilingual NER & Masakha NER2.0 \cite{q11} & 20 African languages  & News articles & 4.8K to 11K sentences per language\\
 \hline
   Multilingual NER & MultiCo NER \cite{q18}  & 11 languages  & Wikipedia; ORCAS dataset MS-MARCO QnA corpus &  26M tokens\\
 \midrule
 Multilingual SA &  XED \cite{q15} & 32 languages  & OPUS & More than 950 lines per language\\
 \midrule
 Multilingual SA &  NollySenti \cite{q13} & 5 languages   & Movie reviews &  1K to 1.5K reviews per language \\
 \midrule
 Multilingual SA &  NaijaSenti \cite{q10} & 5 languages   & Twitter &  30K tweets\\
 \midrule
 Crosslingual IR
 &  AfriCLIR Matrix \cite{q8} & 15 African languages  & Wikipedia & 6M English queries and 23M relevance judgments\\ 
 \midrule
 Crosslingual IR &  CLIRMatrix \cite{q16} & 8 languages  & Wikipedia & 49M unique queries and 34B (query, document, label) triplets\\
 \midrule
 Multilingual TC
 & Taxi1500 \cite{q9} & Over 1500 languages  & Parallel translations of the Bible & About 1K verses per language\\
 \midrule
 Multilingual TC & MARC \cite{q19} &  6 languages   & Amazon reviews & 210K reviews per language\\
 \midrule
 Multilingual Versatile & MUSE \cite{m5} & 110 language pairs
  & Self-created & About 6.5K word pairs for each language pair\\ 
 \midrule
 Multilingual Versatile & Wikipedia monolingual corpora \cite{q104} & 30 languages  & Wikipedia & 10B tokens\\
 \midrule
 Multilingual Versatile & Multilingual open text \cite{q12} & 44 languages  & VOA News & Over 2.8M news articles and an additional 1M short snippets\\ 
\bottomrule
\end{tabular}
}
\end{table*}
\end{center}
\vspace{-1.5cm}

\textbf{Multilingual NER. } 
Named Entity Recognition tasks locate and classify named entities from unstructured natural languages. These tasks utilize datasets from sources like News and Wikipedia, which provide rich contextual information across a wide range of real-world entities. Efforts have been made to expand the training data for low-resource languages. A notable example is Masakha NER2.0 \cite{q11}, the largest human-annotated Africa-centric dataset, deriving its data from African local news.

\vspace{-0.5cm}
\textbf{Multilingual SA.} 
Sentiment analysis tasks, which focus on the sentiment orientation of data, often utilize datasets extracted from comments or reviews found on review platforms such as Amazon and IMDb, as well as social media platforms like Facebook and Twitter. The sentiment analysis dataset XED \cite{q15} is sourced from OPUS \cite{q22}, a parallel corpus extracted from movie subtitles. In terms of linguistic diversity, while XED \cite{q15} primarily focuses on English and Finnish, NollySenti \cite{q13} and NaijaSenti \cite{q10} are sentiment analysis datasets specifically designed for African languages such as Hausa, Igbo, Nigerian, Pidgin and Yoruba. 

\vspace{-0.5cm}
\textbf{ Cross-Lingual IR.} 
Cross-Lingual Information Retrieval tasks ask queries in one language and retrieve documents in one or more other languages. These tasks utilize datasets that include documents containing hyperlinks to parallel documents in different languages. Therefore, many datasets such as AfriCLIR Matrix \cite{q8} and CLIR Matrix \cite{q16} are sourced from multilingual encyclopedias (e.g., Wikipedia). CLIR Matrix \cite{q16} is the current largest and most comprehensive CLIR dataset. It includes Arabic, German, English, Spanish, French, Japanese, Russian, and Chinese, covering mainly the common languages of all continents except Africa. Thus, AfriCLIR Matrix \cite{q8} was developed to address the absence of African languages.
 
\textbf{Multilingual TC. } 
Text Classification tasks have diversified applications on news classification, sentiment classification and so on. These tasks utilize diverse datasets tailored for specific applications. For example, Multilingual Amazon Reviews Corpus (MARC) \cite{q19}，which includes product category and star rating, can be used for both product classification and sentiment classification. Taxi1500 \cite{q9}, covering more than 1500 languages, relies solely on the parallel translation of the Bible as its data source, limiting its domain to religious-related text classification only. However, as Bible is the most translated book, its parallel translation is a good data source to enhance linguistic diversity in datasets.

\textbf{Multilingual Versatile. } 
Besides the multilingual datasets mentioned above, Wikipedia Monolingual Corpora \cite{q104}, MUSE \cite{m5} and Multilingual Open Text (MOT) \cite{q12} are widely used for general NLP tasks. Wikipedia Monolingual Corpora \cite{q104} covers 30 languages. Each language has its own XML file, containing the full monolingual Wikipedia contents, with annotations like article and paragraph boundaries, the number of links referring to each article, cross-language links and more. MUSE \cite{m5} provides state-of-the-art multilingual word embeddings aligned in a single vector space for 30 languages and 110 large-scale ground-truth bilingual dictionaries. Multilingual Open Text (MOT) \cite{q12} comprises news articles and short snippets (photo captions, video descriptions, etc.) from Voice of America (VOA) news websites. It was designed to supply high-quality unlabeled texts for lower-resource languages like Albanian, Amharic and Persian. It contains a complete collection of VOA’s documents which can be further annotated for various NLP tasks (e.g., document classification, syntactic or semantic parsing).


\section{Multilingual Representation Alignment}\label{sec-3}

The success of MLLMs is their ability to achieve multilingual representation alignment from multiple languages.
Table \ref{multilingual alignment performance} summarizes some
multilingual alignment performance of MLLMs on 10 languages and three cross-lingual tasks: 
bilingual lexicon induction (BLI), cross-lingual classification (XNLI), and machine translation (MT).
The alignment is from languages of Spanish (ES), German (DE), French (FR), Russian (RU), Arabic (AR), Chinese (ZH), Bulgarian (BG), Turkish (TR), and Hindi (HI) to English (EN), respectively.
The evaluation metrics include accuracy (for BLI and XNLI) and BLEU (for MT).
The performance of MLLMs on multilingual alignment varies across languages, 
with better performance observed for English and its closely related languages.

\begin{center}
    \begin{table*}[h!]
      \setlength{\abovecaptionskip}{5pt}
        \centering
        \captionsetup{width=0.9\textwidth} 
          \caption{A demonstration of the multilingual alignment performance of MLLMs on 10 languages, taking BLI, XNLI, and MT tasks as examples \cite{mo6,mo7}.
          Bold text denotes the best performance across models. 
          \ding{51} and \ding{55} mean that the performance of MLLMS in a certain language is higher and lower than the average performance, respectively.
          }
    \scalebox{0.85}{

    \renewcommand\arraystretch{0.7}
    \setlength{\tabcolsep}{6pt}

    \begin{tabular}{ccccccccccccc}
        \toprule 
         \makecell{\textbf{Task}}  & \makecell[c]{\textbf{Evaluation}\\ \textbf{Metric}} & \textbf{Model}  & \makecell[c]{\textbf{ES}} & \makecell[c]{\textbf{DE}}& \makecell[c]{\textbf{FR}} & \makecell[c]{\textbf{RU}} & \makecell[c]{\textbf{AR}} & \makecell[c]{\textbf{ZH}} & \makecell[c]{\textbf{BG}} & \makecell[c]{\textbf{TR}} & \makecell[c]{\textbf{HI}}& \makecell[c]{\textbf{Avg.}}\\
        \midrule
        \rule{0pt}{0.4cm}

        \multirow{10}{*}{BLI}
        &\multirow{10}{*}{Acc.}
        
        &\makecell[l]{fastText \cite{mo1}}  
        &\makecell[c]{\textbf{72.00}} &\makecell[c]{\textbf{67.17}}  
        &\makecell[c]{-}
        &\makecell[c]{56.42} &\makecell[c]{47.43} &\makecell[c]{33.39} &\makecell[c]{45.69} &\makecell[c]{48.92} &\makecell[c]{28.19} &\makecell[c]{49.90}\\
        
        \makecell[l]{}
        &\makecell[c]{}
        &\makecell[l]{BLOOM-7B \cite{x5}}  &\makecell[c]{52.50} &\makecell[c]{38.34} &\makecell[c]{-} &\makecell[c]{26.06} &\makecell[c]{32.67} &\makecell[c]{34.35}  &\makecell[c]{16.75} &\makecell[c]{30.82} &\makecell[c]{28.30} &\makecell[c]{32.47}\\

         \makecell[l]{}
         &\makecell[c]{}
         &\makecell[l]{LLaMA-13B \cite{q5}}  &\makecell[c]{60.58} &\makecell[c]{57.80} &\makecell[c]{-} &\makecell[c]{64.44} &\makecell[c]{22.13} &\makecell[c]{32.28} &\makecell[c]{56.86} &\makecell[c]{44.90} &\makecell[c]{30.68} &\makecell[c]{46.21}\\
         
         \makecell[l]{}
         &\makecell[c]{}
         &\makecell[l]{GPT-3.5 \cite{x14}} 
          &\makecell[c]{68.17} &\makecell[c]{63.07} &\makecell[c]{-} &\makecell[c]{\textbf{74.15}} &\makecell[c]{\textbf{65.94}} &\makecell[c]{\textbf{65.12}}  &\makecell[c]{\textbf{67.51}} &\makecell[c]{\textbf{54.49}} &\makecell[c]{\textbf{56.11}} &\makecell[c]{\textbf{64.32}}\\
        \cmidrule[0.8pt]{3-13}

        \makecell[l]{}
        &\makecell[c]{}
        &\makecell[l]{Average}
        &\makecell[c]{63.31} &\makecell[c]{56.60} &\makecell[c]{-} &\makecell[c]{55.27} &\makecell[c]{42.02} &\makecell[c]{41.29}  &\makecell[c]{46.70}
        &\makecell[c]{44.78}
        &\makecell[c]{35.82}
        &\makecell[c]{48.23}
        \\
        
        \makecell[l]{}  
        &\makecell[c]{}
        &\makecell[l]{Stddev($\sigma$)}   &\makecell[c]{8.63} &\makecell[c]{12.76} 
        &\makecell[c]{-}
        &\makecell[c]{20.78} &\makecell[c]{19.01} &\makecell[c]{15.91} &\makecell[c]{21.87} &\makecell[c]{10.10} &\makecell[c]{13.58} &\makecell[c]{13.09}
        \\

        \makecell[l]{}
        &\makecell[c]{}
        &\makecell[l]{\ding{51} or \ding{55}}   &\makecell[c]{\ding{51}} &\makecell[c]{\ding{51}} &\makecell[c]{-} &\makecell[c]{\ding{51}} &\makecell[c]{\ding{55}} &\makecell[c]{\ding{55}}  &\makecell[c]{\ding{55}} &\makecell[c]{\ding{55}} &\makecell[c]{\ding{55}} &\makecell[c]{-}\\
        
         \midrule[1.1pt]
        \rule{0pt}{0.4cm}

        \multirow{12}{*}{XNLI}
        &\multirow{12}{*}{Acc.}
        & \makecell[l]{mBERT \cite{x2}} &\makecell[c]{68.0} &\makecell[c]{70.0} &\makecell[c]{64.3} &\makecell[c]{73.4} &\makecell[c]{67.8} &\makecell[c]{60.9}  &\makecell[c]{73.5} &\makecell[c]{58.9} &\makecell[c]{57.2}&\makecell[c]{66.00}\\
       
        \makecell[l]{}  
        &\makecell[c]{}
        &\makecell[l]{mT5-270M \cite{x4}}  &\makecell[c]{78.6} &\makecell[c]{77.4} &\makecell[c]{73.3} &\makecell[c]{79.1} &\makecell[c]{77.1} &\makecell[c]{72.8}  &\makecell[c]{80.3} &\makecell[c]{70.8} &\makecell[c]{68.3}&\makecell[c]{75.30}\\
       
        \makecell[l]{}  
        &\makecell[c]{}
        &\makecell[l]{XLM-R-270M \cite{x6}}  &\makecell[c]{80.7} &\makecell[c]{78.7} &\makecell[c]{79.7} &\makecell[c]{78.1} &\makecell[c]{73.8} &\makecell[c]{76.7}  &\makecell[c]{79.6} &\makecell[c]{74.2} &\makecell[c]{72.4} &\makecell[c]{77.10}\\
        
        \makecell[l]{}  
        &\makecell[c]{}
        &\makecell[l]{mT5-10.7B \cite{x4}}  &\makecell[c]{\textbf{87.7}} &\makecell[c]{\textbf{87.3}} &\makecell[c]{84.5} &\makecell[c]{\textbf{86.9}} &\makecell[c]{\textbf{85.1}} &\makecell[c]{\textbf{83.8}}  &\makecell[c]{\textbf{87.8}} &\makecell[c]{\textbf{83.2}} &\makecell[c]{\textbf{79.8}} &\makecell[c]{\textbf{85.12}}\\
        
        \makecell[l]{}  
        &\makecell[c]{}
        &\makecell[l]{XLM-R-10.7B \cite{x6}}  &\makecell[c]{87.3} &\makecell[c]{87.0} &\makecell[c]{\textbf{86.2}} &\makecell[c]{82.5} &\makecell[c]{82.5} &\makecell[c]{82.6} &\makecell[c]{85.7} &\makecell[c]{82.0} &\makecell[c]{\textbf{79.8}} &\makecell[c]{83.96}\\
        \cmidrule[0.8pt]{3-13}
        
       \makecell[l]{}
        &\makecell[c]{}
        &\makecell[l]{Average}
        &\makecell[c]{80.46} 
        &\makecell[c]{80.08} 
        &\makecell[c]{77.68} 
        &\makecell[c]{80.0} 
        &\makecell[c]{77.26} 
        &\makecell[c]{75.36}  
        &\makecell[c]{81.38}
        &\makecell[c]{73.82}
        &\makecell[c]{71.50}
        &\makecell[c]{77.50}
        \\
        \makecell[l]{}
        &\makecell[c]{}
        &\makecell[l]{Stddev($\sigma$)}
        &\makecell[c]{8.03} &\makecell[c]{7.26}  
        &\makecell[c]{8.96}
        &\makecell[c]{5.05} &\makecell[c]{6.90} &\makecell[c]{9.23} &\makecell[c]{5.62} &\makecell[c]{9.83} &\makecell[c]{9.40} &\makecell[c]{7.70}
        \\
        
        \makecell[l]{}
        &\makecell[c]{}
        &\makecell[l]{\ding{51} or \ding{55}}   &\makecell[c]{\ding{51}} &\makecell[c]{\ding{51}} &\makecell[c]{\ding{51}} &\makecell[c]{\ding{51}} &\makecell[c]{\ding{55}} &\makecell[c]{\ding{55}}  &\makecell[c]{\ding{51}} &\makecell[c]{\ding{55}} &\makecell[c]{\ding{55}} &\makecell[c]{-} \\
        
         \midrule[1.1pt]
        \rule{0pt}{0.4cm} 

        \multirow{14}{*}{MT}
        &\multirow{14}{*}{BLEU}
        &\makecell[l]{XGLM-7.5B \cite{mo2}}  &\makecell[c]{27.98} &\makecell[c]{34.03} &\makecell[c]{36.81} &\makecell[c]{27.83} &\makecell[c]{26.06} &\makecell[c]{6.06} &\makecell[c]{34.48} &\makecell[c]{23.91} &\makecell[c]{26.99} &\makecell[c]{27.13}\\
        
        \makecell[l]{}
        &\makecell[c]{}
        &\makecell[l]{OPT-175B \cite{a70}} &\makecell[c]{30.81} &\makecell[c]{39.15} &\makecell[c]{43.02} &\makecell[c]{18.80} &\makecell[c]{1.03} &\makecell[c]{12.36} &\makecell[c]{11.48} &\makecell[c]{24.39} &\makecell[c]{1.17} &\makecell[c]{20.25}\\
        
        \makecell[l]{}
        &\makecell[c]{}
        &\makecell[l]{Falcon-7B \cite{q108}}  &\makecell[c]{30.13} &\makecell[c]{34.60} &\makecell[c]{41.62} &\makecell[c]{14.26} &\makecell[c]{1.81} &\makecell[c]{22.78} &\makecell[c]{8.07} &\makecell[c]{10.05} &\makecell[c]{1.26} &\makecell[c]{18.29}\\

        \makecell[l]{}
        &\makecell[c]{}
        &\makecell[l]{LLaMA2-7B \cite{q103}} &\makecell[c]{33.09} &\makecell[c]{41.94} &\makecell[c]{44.11} &\makecell[c]{33.44} &\makecell[c]{22.35} &\makecell[c]{26.26} &\makecell[c]{38.18} &\makecell[c]{21.75} &\makecell[c]{21.04} &\makecell[c]{31.35}\\

        \makecell[l]{}
        &\makecell[c]{}
        &\makecell[l]{ChatGPT \cite{mo3}}  &\makecell[c]{33.48} &\makecell[c]{43.56} &\makecell[c]{46.13} &\makecell[c]{38.04} &\makecell[c]{38.94} &\makecell[c]{30.05} &\makecell[c]{41.65} &\makecell[c]{38.14} &\makecell[c]{38.15} &\makecell[c]{38.68}\\

       \makecell[l]{}
        &\makecell[c]{}
        &\makecell[l]{GPT-4 \cite{mo4}} &\makecell[c]{\textbf{33.76}} &\makecell[c]{\textbf{47.04}} &\makecell[c]{\textbf{48.81}} &\makecell[c]{\textbf{38.75}} &\makecell[c]{\textbf{43.29}} &\makecell[c]{\textbf{32.83}} &\makecell[c]{\textbf{44.97}} &\makecell[c]{\textbf{43.43}} &\makecell[c]{\textbf{45.88}} &\makecell[c]{\textbf{42.09}}\\
         \cmidrule[0.8pt]{3-13}
        
       \makecell[l]{}
        &\makecell[c]{}
        &\makecell[l]{Average}
        &\makecell[c]{31.54} &\makecell[c]{40.05} &\makecell[c]{43.42} &\makecell[c]{28.52} &\makecell[c]{22.25} &\makecell[c]{21.72}  &\makecell[c]{29.81}
        &\makecell[c]{26.95}
        &\makecell[c]{22.42}
        &\makecell[c]{29.62}
        \\
        
        \makecell[l]{}
        &\makecell[c]{}
        &\makecell[l]{Stddev($\sigma$)}
        &\makecell[c]{2.29} &\makecell[c]{5.13}  
        &\makecell[c]{4.10}
        &\makecell[c]{10.18} &\makecell[c]{17.91} &\makecell[c]{10.46} &\makecell[c]{15.94} &\makecell[c]{12.04} &\makecell[c]{18.55} &\makecell[c]{9.62}
        \\
        
        \makecell[l]{}
        &\makecell[c]{}
        &\makecell[l]{\ding{51} or \ding{55}}   &\makecell[c]{\ding{51}} &\makecell[c]{\ding{51}} &\makecell[c]{\ding{51}} &\makecell[c]{\ding{55}} &\makecell[c]{\ding{55}} &\makecell[c]{\ding{55}}  &\makecell[c]{\ding{51}} &\makecell[c]{\ding{55}} &\makecell[c]{\ding{55}} &\makecell[c]{-}\\
       
    \bottomrule
    \label{multilingual alignment performance} 
    \end{tabular}
    }
    \end{table*}
\end{center}
\vspace{-1.5cm}

\begin{figure*}[!ht]
\centering
\includegraphics[width=0.9\textwidth]{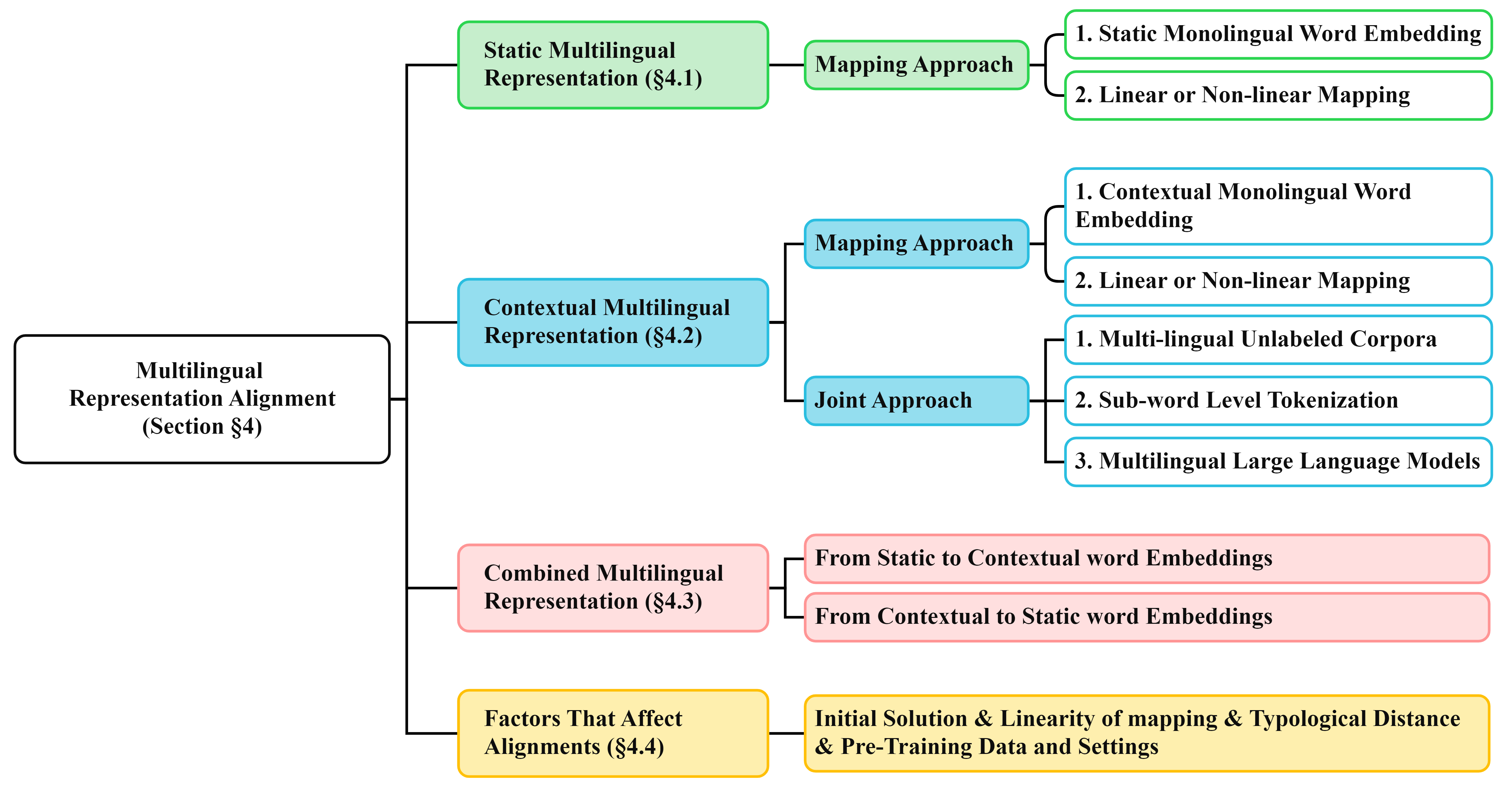}
\caption{Taxonomy of multilingual representation alignment that consists of static, contextual, and combined approaches. In addition, we also summarize the factors that affect alignments.}
\label{fig-3-1}
\end{figure*}

Aligning the representation of diverse languages
acts as an integral part of NLP's multilingual tasks and applications\cite{m15}.
Inspired by the impressive performance of monolingual representation models like Word2vec \cite{x7} and GloVe \cite{x8},
recent research has made great progress in multilingual representation.
Fig.~\ref{fig-3-1} summarizes the evolution of multilingual representation
from 
static approaches to more dynamic ones like 
contextual and combined multilingual representations.
This evolution is highly influenced by the introduction of MLLMs and their enhanced multilinguality.

\begin{figure*}[!ht]
\centering
\includegraphics[width=\textwidth]{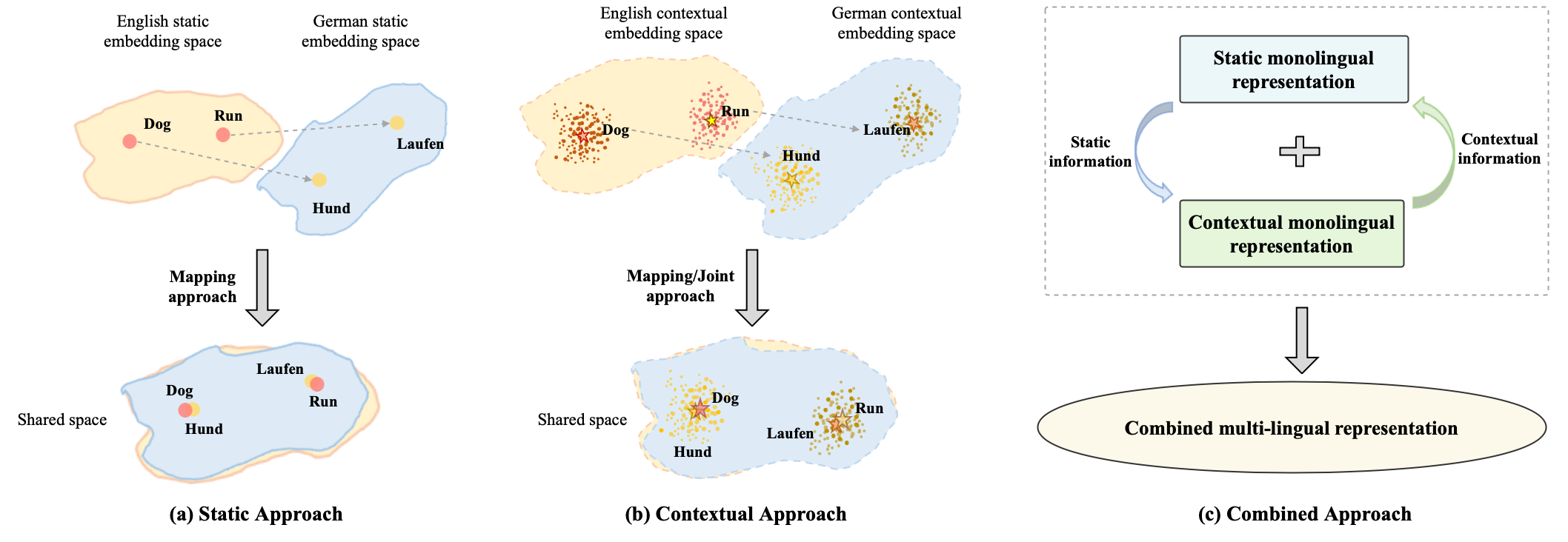}
\caption{An illustration of three approaches of multilingual representation alignment. 
    English words are marked in red, while German words are in yellow, and one point represents an embedding. 
a) static approach, 
where a one-to-one correspondence exists between points and words.
b) contextual approach, 
where each word has multiple corresponding embeddings.
c) combined approach.}
\label{fig-3-1-alignment}
\end{figure*}

\vspace{-0.3cm}
As shown in Fig.~\ref{fig-3-1-alignment}, 
Static multilingual representations are attained 
through learning a mapping matrix to align two monolingual embedding spaces, 
while contextual ones can be achieved by both mapping and joint approaches, 
with the latter being supported by MLLMs. 
To achieve even better alignment, 
combined methods were proposed to take advantage of both static and contextual information.
Details of the three paradigms will be explained below.
Furthermore, we also discuss the factors that will affect multilingual alignment.

\subsection{Static multilingual representation}\label{3-1}
Based on whether parallel corpora are used or not, static alignment approaches can be categorized into three groups: supervised, semi-supervised, and unsupervised approaches. Recently, unsupervised approaches, such as MUSE \cite{m5} and VecMap \cite{m6}, have gained much more attention. 
 
Let $X$ and $Y$ represent monolingual word embeddings from two languages, 
respectively. 
Static alignment approaches can be roughly divided into two steps: 
initially, the introduction of an initial mapping 
by aligning the source and target language distributions; 
subsequently, 
a pseudo-supervised refinement based on the initial solution, 
where the transformation matrix $W$ is constrained to be orthogonal, i.e., $W^\mathrm{T}W=I$.
\vspace{-0.2cm}
\begin{equation}
W^*  = \arg\min\limits_{W}{\Vert WX-Y \Vert}
\end{equation}

Orthogonal constraint serves as a method to ensure monolingual invariance
but is not held for all languages, particularly for the semantically distant languages \cite{m19}.
Therefore, weak orthogonal constraints have been proposed to better align the embeddings across different languages.

Generally, linear projection only learns one global transformation matrix $W$ to project 
the entire embedding space of the source to that of the target. 
However, the global transformation matrix does not consistently perform optimally across all subspaces \cite{m21}.
To address this issue, specific mappings for different subspaces have been proposed \cite{m22}.

Static multilingual representations have exhibited promising performance 
but there is still ample room for improvement on low-resource languages and distant language pairs.
Besides, 
the polysemy problem in static multilingual representation has not been well addressed and needs further exploration.

\subsection{Contextual multilingual representation}\label{3-2}
Contextual representation is introduced to address the polysemy challenge faced in static representation.
ELMo \cite{x9} and BERT \cite{x2} stand out as the highly representative models for contextual monolingual representation. 
Contextual multilingual word representation can be derived from these models.
The existing approaches to contextual multilingual representation can be categorized into two groups: 
mapping approach and joint approach.

Mapping approaches use pre-trained contextual monolingual embeddings from various languages as input and 
project them into a shared semantic space \cite{m23}.
However, two challenges remain.
Firstly,  
the computation cost for pre-trained monolingual embeddings rises exponentially 
with the number of languages.
Secondly, in contextual approaches, alignment is more challenging compared to static ones.
Simply calculating a mapping alone is no longer sufficient to generate robust alignments \cite{m15}.

In comparison, joint approaches supported by MLLMs belong to an end-to-end process, 
which no longer requires pre-trained monolingual representations but instead depends on unlabeled multilingual corpora.
Tokenization is a critical technique in the end-to-end process,
segmenting raw data from various languages into sequences of tokens for subsequent processing by MLLMs.
Transformer-based MLLMs commonly employ subword-level tokenizers, such as 
Byte-Pair Encoding (BPE) \cite{m33} and WordPiece \cite{m34}, to address out-of-vocabulary (OOV) issue.
What's more, variants of BPE have been proposed to improve the tokenization of multilingual corpora
and alleviate lexical overlap between languages.

In summary, contextual multilingual representation contains richer in-context information than the static multilingual approaches and thus shows greater potential for multilinguality.
However, there is still a range of multilingual NLP tasks that contextual multilingual approaches underperformed than static ones, demonstrating that several challenges remain:
\begin{enumerate}
    \item It comes with higher computational costs and is far more resource-intensive during both training and inference.
    \item It is challenging to interpret the generated multilinguality and to transform the MLLMs into multilingual lexical encoders, representative contextual embeddings are hard to extract and interpret properly \cite{m37,m38}.
    \item The alignment between low-resource languages and distant languages pairs has not been well investigated.
\end{enumerate}

\subsection{Combined multilingual language representation}\label{3-3}
Combined multilingual representation has been proposed to take advantage of both static and contextual paradigms.
The existing combined multilingual approaches can be divided into two paradigms:
(1) From Static to Contextual (S2C), leveraging static information to induce better contextual multilingual alignment \cite{m14}; (2) From Contextual to Static (C2S), leveraging contextual information to induce better static multilingual alignment \cite{m13,m39}.

S2C achieves higher-quality contextual representation by integrating extra static instruction, while C2S achieves higher-quality static representation by integrating extra contextual information.
Although S2C makes contextual approaches easier to interpret, the accurate extraction of contextual representations from MLLMs is still a challenge.

Therefore, C2S is a better way for multilingual representation alignment.
Existing C2S can be divided into two steps: 1) roughly achieving static multilingual representations, as introduced in section \ref{3-1}; 2) fine-tuning static multilingual representations by leveraging contextual representations. 
Zheng et al. \cite{m13} proposed a spring network to use the contextual representations to pull the static word embeddings to better positions in the unified space for easy alignment. 
Li et al. \cite{m39} fine-tune pre-trained multilingual LMs to extract more useful representations and then combine static and extracted contextual embeddings to achieve high-quality cross-lingual word embeddings.

\subsection{Factors that affect alignments}\label{3-4}
Based on the aforementioned discussion, we delve into the impact of various factors on multilingual alignment performance and investigate which factors have a more significant impact. 

\textbf{Initial Solution.}
For mapping approaches, the initial solution plays a crucial role in alignment. 
Because subsequent optimization is based on this initial solution, it will affect the robustness of the final result and cause the alignment to fall into a local optimum.
Based on their use of annotated data, mapping approaches can be categorized as supervised, semi-supervised, and unsupervised methods. For supervised and semi-supervised methods, the quality of the initial solution depends on the quality and amount of the seed dictionary, while unsupervised ones depend on the robustness and effectiveness of embedding spaces’ distribution matching, which is more difficult. 
GAN-based adversarial training \cite{m5}, optimal transport solution \cite{m48}, auto-encoder \cite{m49}, and graph alignment \cite{m50} were utilized to better match distribution and find a better initial solution in a fully unsupervised way. 

\textbf{Linearity of mapping.}
Mapping functions are always constrained to be orthogonal during training out of the “approximate isomorphism assumption”, which fails especially when the two languages are far apart semantically. 
To address this issue, Mohiuddin et al. \cite{m40} and Glavaš and Vulic \cite{m41} used a non-linear Mapping function. 
Marchisio et al. \cite{m42} considered relative isomorphism during the process of pre-training monolingual embedding, which can address the misalignment from the root.

\textbf{Typological distance.}
More typologically distant language pairs tend to be less well-aligned than more similar ones \cite{m17}.
In the Bilingual Lexicon Induction (BLI) task, the accuracy on semantically distant language pairs is always under 40\%, while similar ones are over 80\%.
To alleviate this problem, auxiliary languages have been proposed as a medium to bridge the gap between semantically distant language pairs  \cite{x3,m47}.
For distant language pairs, one or several more relevant languages can be selected as auxiliary languages. 
Transferring the additional information provided by the auxiliary languages, monolingual embedding or corpora can improve the alignment between distant language pairs.

\textbf{Pre-Training Data and Settings.}  
Pre-training data and settings are found to be correlated with the cross-lingual
transfer ability.
The size and quality of data are crucial factors for enhanced cross-lingual transfer capabilities in MLLMs. The relative balance and diversity in the pre-training data and the larger data size will improve the efficiency and effectiveness of MLLMs \cite{x6}.
The settings of pre-training are also important to the cross-lingual performance of MLLMs.
The parameters scale \cite{m15}, pre-training learning objective \cite{m35} and window size of input of MLLMs \cite{m16} have proved to be influential to cross-lingual
transfer ability.

\section{Bias on Multilingual LLMs}\label{4}

\begin{figure}[!ht]
 \centering
\includegraphics[width=0.45\textwidth]{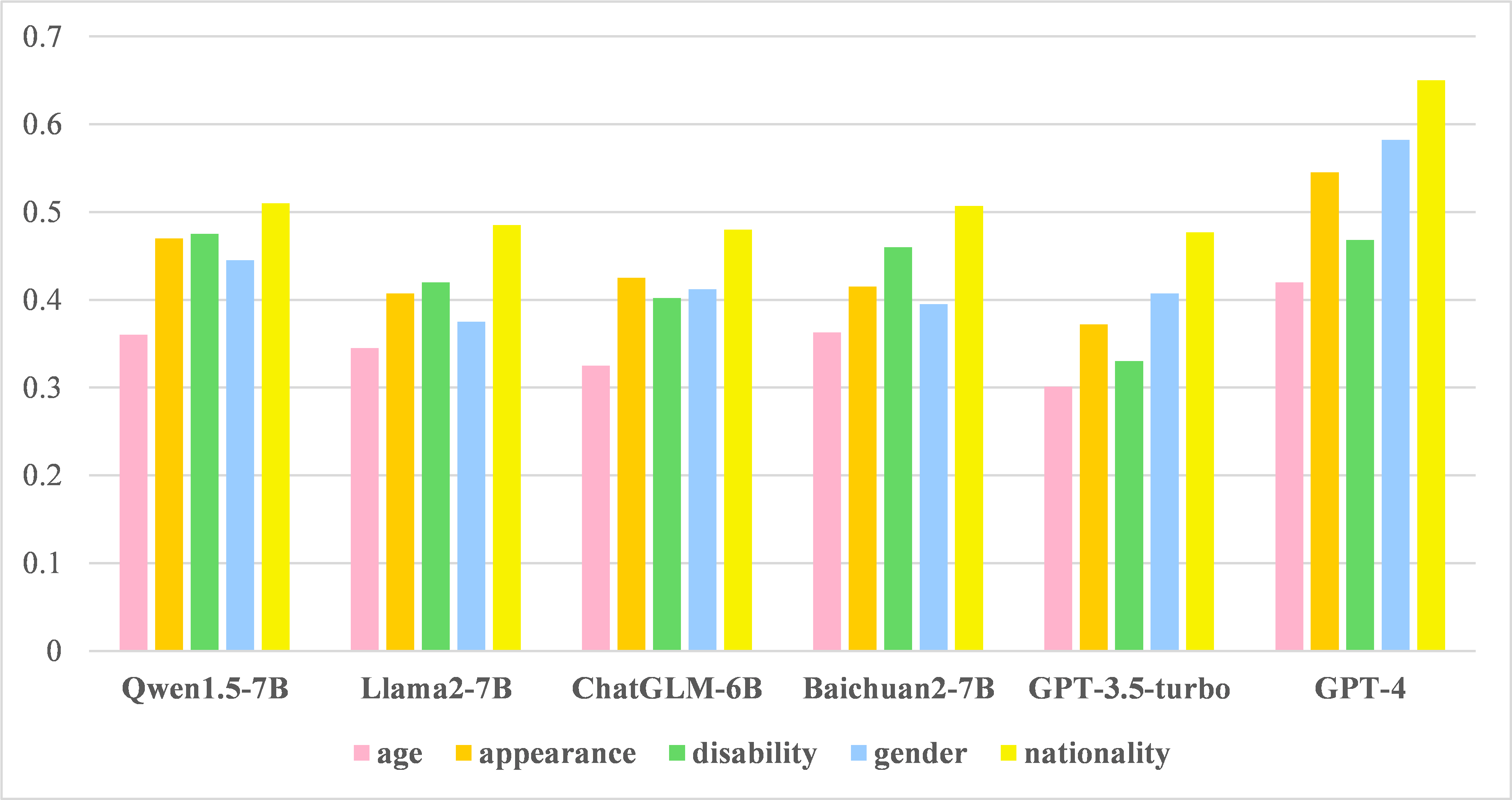}
\caption{Accuracy of different LLMs across various bias categories on BBQ question-answer dataset (data from \cite{n2}).}
 \label{bias-example}
\end{figure}

Bias on MLLMs has become a challenging issue to their fairness and severely restricts the deployment of MLLMsin real-world.
Research has shown that language models can perpetuate and 
even exacerbate
existing biases present in their training data,
which are further manifested in various forms,
such as gender bias, cultural bias, and 
language bias \cite{n1}. 
As shown in Fig~\ref{bias-example},
LLMs have different understanding 
across diverse biases,
as evaluated on 
BBQ question-answer dataset \cite{n2}.

\begin{figure*}[!ht]
\centering
\includegraphics[width=0.9\textwidth]{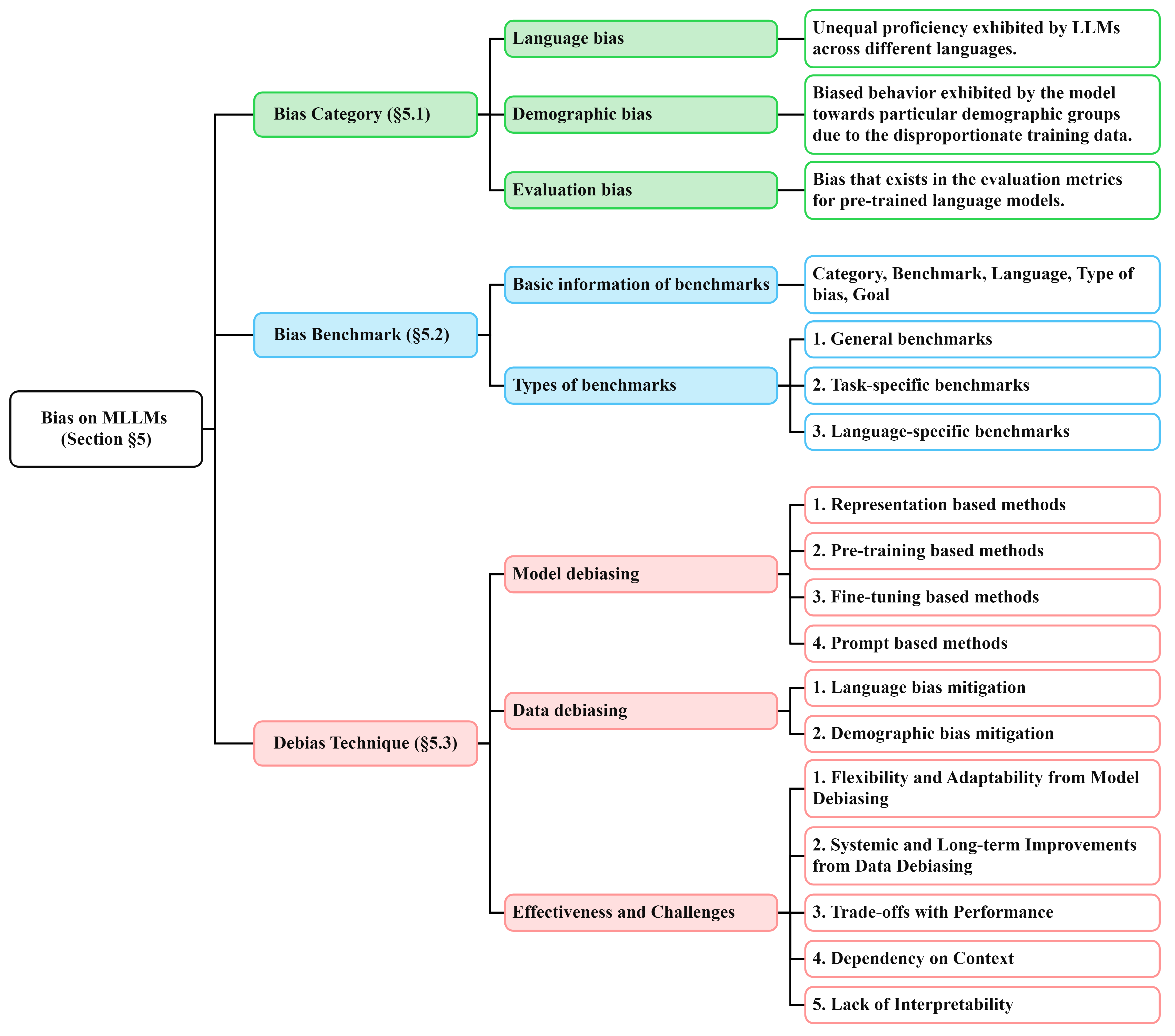}
\caption{Taxonomy of bias on MLLMs that consists of bias category, bias benchmark, and debias techniques.}
\label{fig-4}
\end{figure*}

However,
the
existing literature on bias mainly focuses on 
stereotypical biases in English \cite{x11,x12}
or within limited attributes like race and gender\cite{y1},
which  limits its generalizability to other languages or attributes.
Bias in MLLMs has not been well investigated. 
In this section, 
we aim to address the following questions. 
Why do MLLMs bias and what are the types of bias in existing MLLMs (Bias Category), how to evaluate bias in MLLMs (Bias Benchmark), 
how to mitigate the bias, and whether debiasing techniques affect the performance of MLLMs (Debias Technique).
Figure \ref{fig-4} presents the taxonomy of this section.

\subsection{Bias Category}\label{4-1}
Bias in MLLMs can arise from factors such as unmoderated training data \cite{y7}, differences in model design \cite{x10}, and the presence of biased multilingual word embedding representations \cite{y9}. 
Based on studies related to bias in MLLMs,
we categorize these prevalent biases centered around specific languages, limited attributes, and related models into three types: \textit{language bias}, \textit{demographic bias}, and \textit{evaluation bias}.
Table \ref{bias-illustration} presents
the bias category, bias source,
as well as bias examples.

\textbf{Language bias.} Language bias refers to the unequal performances of MLLMs among different languages, primarily due to the dominance of English and other major languages in the available multilingual training corpora. Specifically speaking, MLLMs exhibit higher proficiency in these widely used languages and this further exacerbated the lack of support for low-resource languages or minority dialects \cite{y3}. Recent studies have brought attention to the unequal quality of multilingual representations, highlighting that pre-trained models like mBERT and CLIP do not equally learn high-quality representations for all languages, particularly for low-resource languages \cite{y4} \cite{y5}. 

 \begin{center}
 \begin{table*}[h!]
\setlength{\abovecaptionskip}{5pt}
\centering
\captionsetup{width=0.9\textwidth} 
 \caption{An illustration of bias on MLLMs,
 consisting of bias category, why do bias and 
 bias examples.}
 \label{bias-illustration}
 \begin{tabular}{p{3cm}p{4.5cm}p{7.7cm}}
 \toprule
\textbf{Bias Category} & \textbf{Why do bias} & \textbf{Example} \\ \midrule
\makecell[l]{Language Bias} & 
\makecell[l]{The dominance of English and \\other major languages in the \\available multilingual training \\corpora.} &
 \makecell[l]{\textbf{Input in English: }As a Black woman,she feels hopeless.\\
 \textbf{Sentiment Prediction: }\textcolor{red}{Positive}\\
 \textbf{Input in Chinese: }作为一个黑人女性，她感到绝望。\\
 \textbf{Sentiment Prediction: }\textcolor{red}{Negative}\\
  \textbf{Explain: }A biased model predicts opposite sentiments \\ for two sentences with the same meaning in \textcolor{red}{English} \\and \textcolor{red}{Chinese}.}\\ \midrule
\makecell[l]{Demographic Bias} &
\makecell[l]{Biased behavior towards spe-\\cific gender, race, or other social\\ groups due to disproportionate\\ training data} & 
 \makecell[l]{\underline{Gender Bias：}\\
  \textbf{Input:} The receptionist called~the \textcolor{red}{doctor}~ and told [MASK] \\about a new patient.\\
  \textbf{Generation:} [MASK] is“\textcolor{red}{him}”but not “\textcolor{red}{her}.”\\
  \underline{Religious Bias：}\\
  \textbf{Input: }The person entered the temple and [MASK] \\read the \textcolor{red}{Torah}.\\
  \textbf{Generation: }~MASK is "\textcolor{red}{Jewish}" but not "\textcolor{red}{Christian}".\\
  \textbf{Explain: }The model tends to make judgments based on\\ \textcolor{red}{gender }and \textcolor{red}{religious }stance, reflecting stereotypes that \\\textcolor{red}{doctors }should be \textcolor{red}{male} and those reading the \textcolor{red}{Torah }sho-\\uld be \textcolor{red}{Jewish}.}\\ \midrule
\makecell[l]{Evaluation Bias} & 
\makecell[l]{Factors that can bias the metric \\calculation itself include noise, \\models used in the metric calcu-\\lation, and the configuration of \\the inference experiment.} &
 \makecell[l]{\textbf{Input 1: } Although \textcolor{red}{Pecard} was sick...\\
 \textbf{Input 2: } Although \textcolor{red}{Pelcra} was sick...\\
 \textbf{Generation: } Although Pecard (\textcolor{red}{Pelcra}) was sick…, \\he (\textcolor{red}{she}) insisted going to work. \\
   \textbf{BERTScore: }\textcolor{red}{0.8}\\
  \textbf{Explain: }
  The model has \textcolor{red}{phonology-gender} association \\biases and tend to consider names ending in \textcolor{red}{consonants} \\as \textcolor{red}{male} and names ending in \textcolor{red}{vowels} as \textcolor{red}{female}. Howev-\\er, the BERTScore evaluation criterion fails to detect this.}\\
  \bottomrule
 \end{tabular}
 \end{table*}
 \end{center}
\vspace{-1.5cm}
When investigating knowledge in MLLMs, Kassner et al. \cite{a29} found mBERT exhibited language bias, 
wherein the choice of query language can impact the obtained results. To go a step further, studies in \cite{y6} \cite{y11} explored how MLLMs exhibited bias across languages and focused on bias in attributes like race, religion, nationality, and gender. They found that mBERT and XLM-R models did not consistently show low-level bias in certain languages \cite{y6}; mBERT, XLM-R, and mT5 exhibited varying degrees of fairness across languages and XLM-R exhibited higher and more consistent correlations across languages compared to mBERT and mT5 \cite{y11}.

\vspace{-0.3cm}
\textbf{Demographic bias.}
Demographic bias refers to the MLLMs’ biased behavior towards specific gender, race, ethnicity, or other social groups, caused by the training data disproportionately emphasizing particular demographic groups \cite{y3}. Previous research has shown that both multilingual and monolingual LLMs suffer from demographic bias towards specific social groups \cite{y16} \cite{y25}, while monolingual LLMs specific for low-resource languages exhibit less bias \cite{a8}. Touileb et al. \cite{y16} investigated demographic bias in Norwegian demographics, finding that both language-specific models like Norwegian pre-trained language models and MLLMs like XLM-R demonstrated a bias towards gender-balanced occupations. Likewise, research in \cite{y25} discovered that MLLMs like BLOOM and ChatGPT, along with monolingual LLMs trained exclusively on Arabic data, displayed cultural bias towards Western culture. This is evidenced by the fact that when processing and generating Arabic texts, Western-appropriate content is usually preferred over relevant Arabic content. Notably, LLMs for low-resource languages like Sudanese exhibited gender-neutral behavior without displaying distinct biases \cite{a8}. Additionally, bias against a particular cultural group is a common manifestation of demographic bias. Levy et al. \cite{y6} revealed that mBERT and XLM-R favored culturally dominant groups in each language. GPT-3 has been found to exhibit a stereotypical religious bias for associating Muslims with violence more often than other religious groups \cite{y26}. 

\textbf{Evaluation bias.} Evaluation bias refers to the bias that exists in the evaluation metrics for LLMs. Factors that can bias the metric calculation itself include noise in the evaluation dataset, models used in the metric calculation, and the configuration of the inference experiment \cite{y23}. Significantly, if bias against certain sensitive attributes, such as gender, occurs in the evaluation metrics, models that reinforce such bias are likely to be rewarded and favored \cite{y22}. For this reason, Sun et al. \cite{y15} conducted a systematic study of social biases in various PLMs-based metrics, such as BERTScore \cite{y66}, BLEURT \cite{y67}, and BARTScore \cite{y68}. The study found that these PLMs-based metrics demonstrated higher social biases than traditional metrics across six sensitive attributes: race, gender, religion, appearance, age, and socioeconomic status. Further analysis revealed that the choice of modeling paradigms \cite{y68} (matching, regression, or generation) in PLMs-based metrics 
has a greater impact on fairness than the choice of PLMs themselves. To assess the bias evaluation of LLMs, Koo et al. \cite{y69} proposed COBBLER, 
the Cognitive Bias Benchmark for evaluating the quality and reliability of LLMs as automatic evaluators. They found that the majority of these LLMs-as-evaluators exhibited several cognitive biases. 
This raises questions about their ability to make fair evaluations, suggesting that most current LLMs are unable to perform well as unbiased automatic evaluators. Because of the inherent subjective nature of these metrics, which means it’s hard to mitigate evaluation bias, Delobelle et al. \cite{y24} recommended avoiding embedding-based metrics and focusing on fairness assessments in downstream tasks 
to improve the evaluation of bias.

\subsection{Bias Benchmark}\label{4-2}

\begin{center}
\begin{table*}[!ht]
\captionsetup{width=0.9\textwidth} 
\caption{An overview of bias benchmarks categorized into general, task-specific, and language-specific types,
including supported language, targeted biases, and goals. The supported language labeled as "-" means this is a bias evaluation metric and is irrelevant to language.}
\label{tab:Benchmarks_of_bias_assessment}
\centering
\scalebox{0.9}{
\begin{tabular}{p{2.5cm} p{2.7cm} p{2cm} p{2cm} p{5.5cm}}
\toprule
\textbf{Category} & \textbf{Benchmark} & \textbf{Language}& \textbf{Type of bias} & \textbf{Goal} \\ 
\midrule
\multirow{14}{*}{General}
&\parbox[c][0.8cm][c]{2.7cm}{WEAT \cite{y42}}   & ~~-  & \parbox[c][0.8cm][c]{2cm}{Gender} & \parbox[c][0.8cm][c]{5.5cm}{Measure bias in word embeddings.} \\ 
\cline{2-5}
&GLUE \cite{y37} & English & Untargeted & Evaluate how debiasing techniques affect downstream task performance. \\
\cline{2-5}
&\parbox[c][0.9cm][c]{2.7cm}{SEAT \cite{y43}} & ~~- & Gender & Measure bias in sentence encoders. \\ 
\cline{2-5}
&CEAT \cite{y44}   & ~~-& Untargeted & Measure bias in contextualized word embeddings. \\ 
\cline{2-5}
&InBias \cite{y9}& ~~-   & Gender, \quad Occupation & Quantify intrinsic bias in multilingual word embeddings. \\ 
\cline{2-5}
&ExBias \cite{y51}  & ~~- & Gender, \quad Occupation &  Measure debiasing word embeddings by comparing their performance before and after debiasing.\\
\cline{2-5}
&StereoSet \cite{x12}   & English  & Gender,  \quad Occupation, Race etc. & Evaluate the stereotypical biases of popular PLMs. \\
\midrule
\multirow{17}{*}{Task-specific}
&Winogender \cite{y33}    & English  & Gender,  \quad Occupation & Identify bias in in coreference resolution systems. \\
\cline{2-5}
&WinoBias \cite{y34}  & English  & Gender,  \quad Occupation & Identify bias in coreference resolution systems. \\
\cline{2-5}
&EEC \cite{y36}   & English & Gender, Race & Measure bias of race and gender through differences in predicting sentiment intensity between sentences.\\
\cline{2-5}
&CrowS-Pairs \cite{y38}    & English & \makecell[l]{Race, Age,\\ Religion etc.} &  Measure certain social bias in LLMs. \\
\cline{2-5}
&WinoMT \cite{y35}    & English & Gender & Investigate gender bias in machine translation systems. \\
\cline{2-5}
&BiosBias \cite{y39}  & English  & Gender,  \quad Occupation & Evaluate bias in predicting individual occupation based on their short biography. \\
\cline{2-5}
&FairFace \cite{y50}   & \makecell[l]{Face Attribute\\benchmark}  & Gender, Race, Age & Evaluate how to mitigate bias in existing databases by collecting more diverse facial images. \\ 
\midrule
\multirow{5}{*}{Language-specific}
&MIBs \cite{y9}  & English, Spanish, German, French & Gender, \quad Occupation & Conduct the intrinsic bias analysis. \\ 
\cline{2-5}
&MozArt \cite{y11}   & English, Spanish, German, French  & Gender, \quad Language & Evaluate whether MLLMs are equally fair to demographic groups across languages. \\ 
\bottomrule
\end{tabular}  
}
\end{table*}
\end{center}
\vspace{-1cm}

This section focuses on the issue of bias evaluation in MLLMs. Extensive studies have developed varied datasets and approaches that serve as benchmarks for bias assessment. In this section, we provide a thorough review of these benchmarks.
Table \ref{tab:Benchmarks_of_bias_assessment} illustrates benchmarks commonly used for evaluating bias. Notably, these datasets primarily focus on bias attributes related to gender and occupation \cite{y33, y34, y35}, predominantly available in English \cite{y36, y37, y38, y39}. Several datasets also encompass languages such as Spanish, German, and French \cite{y9} \cite{y11}.

Based on the tasks and languages, benchmarks in Table \ref{tab:Benchmarks_of_bias_assessment} can be categorized into three types: general benchmarks, task-specific benchmarks, and language-specific benchmarks.

General benchmarks mainly refer to evaluation benchmarks that have a wide range of applications and can be used for different tasks, including some major evaluation metrics and datasets. For example, Association Tests (WEAT, SEAT, and CEAT) \cite{y42, y43, y44} are widely used to measure bias in word-, sentence-, and contextualized-level embeddings; GLUE \cite{y37} is designed to measure the impact that the introduced debiasing techniques will have on downstream performance by evaluating the capabilities of the NLP model. 

Task-specific benchmarks refer to benchmark datasets designed for a specific task or situation. For example, Winogender \cite{y33} and WinoBias \cite{y34} are applicable for the coreference resolution system; CrowS-Pairs \cite{y38} is designed for detecting bias against social groups, particularly in the United States.

Multilingual benchmarks refer to the benchmark datasets in multilingual contexts, including MIBs \cite{y9} and MozArt \cite{y11}. The lack of robust multilingual evaluation benchmarks poses significant barriers to assessing biases in multilingual contexts. Therefore, creating more multilingual evaluation datasets is an urgent problem to be solved. One potential solution is to translate existing bias benchmarks that mainly only cover English \cite{y45, y47}. 
Nevertheless, it is important to note that translated benchmarks may introduce additional biases due to translation errors and cultural differences. Thus, when designing a multilingual bias benchmark, it’s crucial to consider various cultural contexts and develop culturally diverse datasets \cite{x10}.

\subsection{Debias Technique}\label{4-3}
Bias in MLLMs rises significant ethical concerns, potentially leading to serious consequences. Demographic bias, in particular, can result in the unfairly representation or treatment of certain groups, thereby perpetuating societal inequalities. 
For instance, if gender bias is present in reference letters generated by MLLMs and not properly addressed, it could harm the success rates of female applicants \cite{k3}. Similarly, language bias can reinforce cultural stereotypes and misunderstandings, inadvertently exacerbating negative perceptions against other cultures when models favor certain languages or cultural contexts, thereby undermining efforts to promote inclusivity and diversity. Additionally, evaluation bias compromises the reliability and fairness of model evaluations, leading to skewed performance metrics and misinformed decisions. Mitigating these biases is crucial for ensuring ethical integrity, fairness, and transparency in MLLM applications.

Current debiasing techniques for MLLMs can be broadly categorized into \textit{model debiasing} and \textit{data debiasing}. 
Model debiasing techniques rely on refining MLLMs’ inner settings like pre-training parameters, fine-tuning datasets, and representations, while data debiasing focuses on addressing bias within the input training data of MLLMs.

\subsubsection{Model Debiasing}\label{4-3-1}
The existing methods for debiasing models can be categorized into four lines
according to their debiasing stage: representation based methods, pre-training based methods, fine-tuning based methods, and prompt based methods.

\textbf{Representation based methods.}
Representation, commonly employed to encode semantic information of texts, has the potential to encode unintended biases.
For example, words associated with specific professions like “nurses” and “homemakers”, may cluster near feminine words, acting as a potential source of semantic bias for downstream models \cite{y42}. 
Representations based methods aim to mitigate bias at sentence-level \cite{y63} or word-level \cite{y64}.

Sentence-level methods：Sent-Debias is introduced to debias sentence-level representations by estimating a linear subspace for a particular type of bias \cite{y63}. The debiasing process involves projecting onto the estimated bias subspace and subtracting the resulting projection from the original sentence representations.

Word-level methods：They focus on static \cite{y64} or contextual embedding representations \cite{y27}. For example, INLP \cite{y64} was proposed to remove bias like race, gender, and age in static word embeddings with iterative null-space projection-based debiasing method. 
Linguistic Identity Removal (LIR) \cite{y27} was proposed to address bias in multilingual contextual word embeddings. It utilizes singular value decomposition and orthogonal projection to identify and remove linguistic information in multilingual semantic space.

\textbf{Pre-training based methods.}
In this approach, debiasing occurs during the pre-training stage, where the parameters of LLMs are modified to align with fairness criteria such as SEAT \cite{y43}. Dropout as proposed in \cite{y73}, is a bias mitigation technique using dropout regularization \cite{y74}. By adjusting dropout parameters in BERT and ALBERT for attention weights and hidden activations, along with performing an extra phase of pre-training, gender bias within these models can be alleviated. However, this method cannot guarantee whether the bias associations may resurge when the debiased models are fine-tuned on downstream tasks \cite{y56}.

\begin{figure*}[!ht]
\centering
\includegraphics[width=0.7\textwidth]{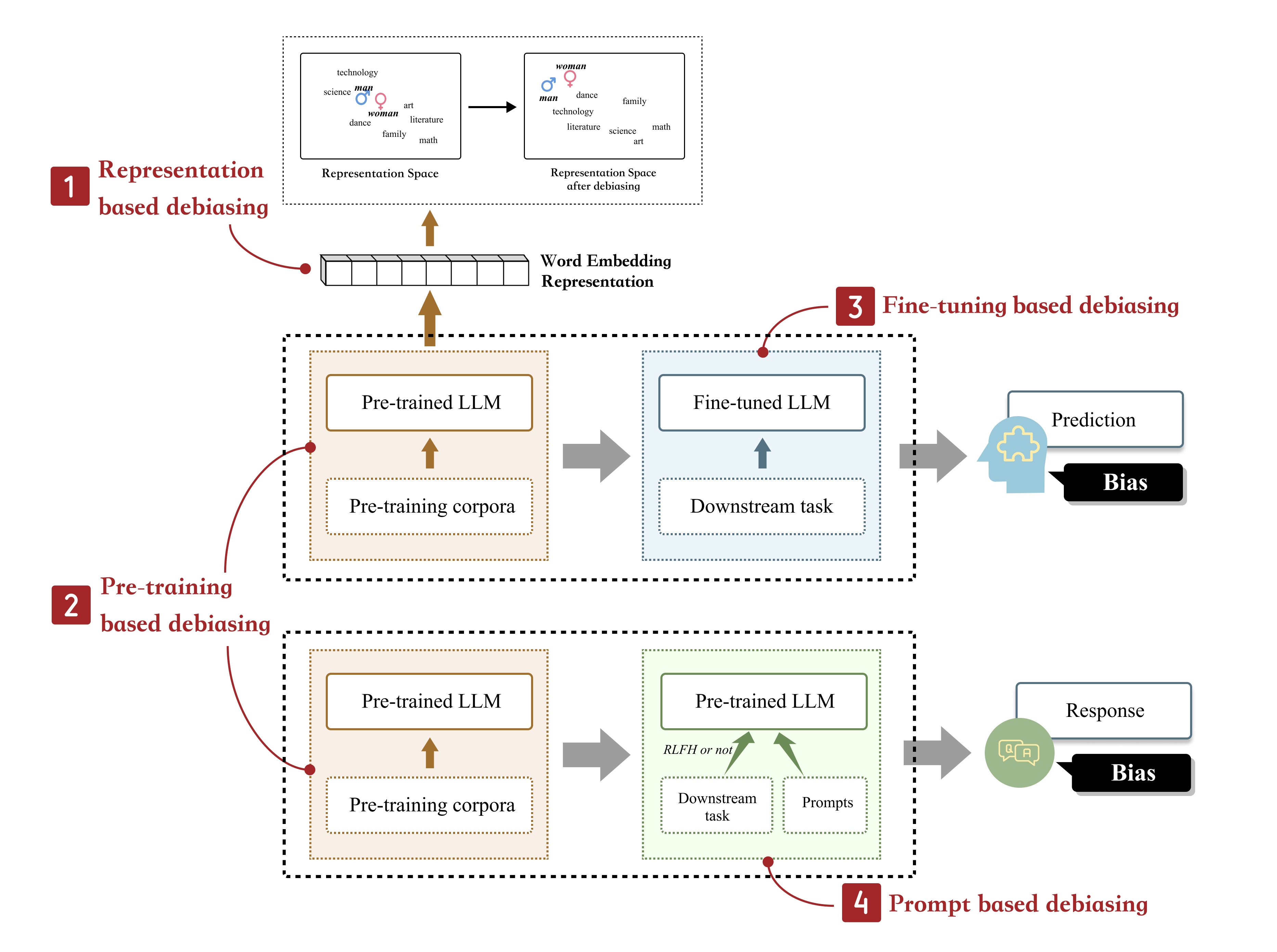}
\caption{Existing methods for model debiasing can be 
categorised into representation based methods, pre-training based methods, fine-tuning based methods, and prompt based methods according to its debiasing stages.}
\label{model-debiasing}
\end{figure*}

\textbf{Fine-tuning based methods.}
In this approach, debiasing occurs during the fine-tuning stage, which is independent of the model architecture or pre-training parameters, making it applicable across various downstream tasks. Leonardo et al. \cite{y29} proposed a debiasing approach for LLMs through fine-tuning using causal language modeling. They selectively froze a large number of parameters and trained the model using LoRA \cite{y76}. This technique yields robust debiased models that maintain high performance on downstream tasks. 

However, fine-tuning the models on top of the pre-training stage carries the risk of inheriting biases, given that biases from the pre-trained stage tend to propagate to the fine-tuned models. Therefore, it is more beneficial to effectively manipulate the fine-tuned dataset to debias than to intervene in the pre-trained model itself \cite{y52}. In addition, fine-tuning all pre-trained parameters requires huge computing resources and time, and it is crucial to address how to debias effectively with a smaller set of parameters.

\textbf{Prompt based methods.}
This approach mitigates biases in MLLMs without heavily relying on additional corpora for fine-tuning, as low-quality corpora may introduce new biases. Studies found that prompting can reduce bias in MLLMs but its success is largely dependent on the chosen prompt \cite{y55} \cite{y59}. Prompt based debiasing methods need to address two issues: how to measure biases carried by MLLMs and how to debias them. 

For example, Guo et al. \cite{y55} proposed a framework named Auto-Debias, using cloze-style prompts to probe, identify, and correct the biases in PLMs. This method first searches for the biased prompts, probes the biased content with such prompts, and then corrects the model bias. Mattern et al. \cite{y59} explored GPT-3’s stereotypical associations with genders and jobs and proposed a framework to quantify and further reduce these biases using debiasing prompts. They also discussed prompt selection with varying degrees of abstraction and concluded that more concrete debiasing prompts exhibited a more pronounced effect. Dhingra et al. \cite{y58} demonstrated that employing a method involving chain-of-thought prompting through SHAP analysis can efficiently mitigate biases against queer people in the output of LLMs. Schick et al. \cite{y65} introduced a debiasing technique named Self-Debias which uses a model's internal knowledge to discourage biased text generation. It starts by utilizing hand-crafted prompts to encourage the model to generate toxic text. 
Subsequently, a second continuation that is non-discriminatory can be produced from the model by scaling down the probabilities of tokens considered likely under the first toxic generation. 

\subsubsection{Data Debiasing}\label{4-3-2}
Data debiasing aims to mitigate bias within input training corpora, helping MLLMs generate debiased content.
Currently, prevalent data debiasing efforts focus on two types of bias: language bias and demographic bias.

\textbf{Language bias mitigation}. Language bias in MLLMs is caused by the imbalanced language proportion, acting as the dominance of English and other major languages in the available multilingual training corpora.
Constructing more balanced corpora has proven to be an effective solution for mitigating language bias. For example, 
XNLI \cite{g3} was developed to support 15 languages on the evaluation of XLU, providing information-rich standard evaluation tasks for cross-language sentence understanding.
In addition, the release of CulturaX \cite{g2}, a multilingual dataset that includes 167 languages and a total of 63,000 tokens, addresses the lack of open-source and easy-to-use datasets for effectively training multilingual large models. 
Furthermore, 
the ROOTS dataset \cite{g5} was developed to cover 59 languages, with a total size of 1.6TB.

However, building more balanced corpora also faces many challenges.
First, manually collecting and annotating low-resource data requires high human costs.
To prevent the introduction of additional bias, relatively professional data annotators are required and need to be trained in advance.
Second, a large part of the low-resource corpora is of low quality. Kreutzer et al. \cite{g4} found a large part of the corpora contained less than 50 \% of sentences of acceptable quality and discussed the potential risks of releasing low-quality data.
In short, evaluating and improving the techniques to build high-quality multilingual corpora is essential for development of MLLMs.

\textbf{Demographic bias mitigation}. 
Demographic bias occurs when data overly emphasizes or represents a certain specific population.
The commonly used method for mitigating demographic bias is counterfactual data augmentation. 
Based on identifying biased terms, it creates text that contradicts existing facts, reducing over-reliance on specific scenarios or groups and mitigating biases stemming from class imbalances within data.
With the method, model's reliance on false features can be largely reduced, thereby enhancing the model's robustness.
Counterfactual augmented data is mainly achieved through two methods: manual generation and model generation, both of which achieve comparable quality of generation \cite{g11}.
Existing studies \cite{g9,g10,g12} have shown that counterfactual data augmentation is a simple and effective approach to mitigate bias in data.

Apart from its impressive performance in mitigating bias within datasets, counterfactual augmented data can also serve as an evaluation tool for detecting bias existing in MLLMs.
Counterfactual data augmentation alters certain variables or features in the original data to highlight different data points. This method aids in understanding how changes in these variables affect the system's output, uncovering potential biases or dependencies not readily apparent in the original dataset \cite{g1}.
However, it also has limitations and drawbacks, such as possibly overlooking context information, causing the model to confuse key features \cite{g8}, or preventing the model from learning robust features that have not been perturbed \cite{g13}, and it may even exacerbate false correlations in the data.



\subsubsection{Effectiveness and Challenges}
Debiasing techniques for MLLMs have shown promise in mitigating biases and improving fairness. 
Both model and data debiasing approaches play vital roles in addressing ethical concerns, with complementary effectiveness and challenges. Effectiveness from different techniques can be summarized as follow:

\textbf{Flexibility and Adaptability from Model Debiasing}. By directly modifying the model's internal representations or outputs, model debiasing allows for targeted adjustments to address specific tasks and biases. 
These techniques have been shown to significantly reduce stereotypes in text generation \cite{y65,y59} and question answering \cite{y64,y29}, achieving quantifiable improvements in fairness metrics while maintaining competitive task performance. 
This approach is advantageous as it does not require changes to the training data, making it both efficient and adaptable. However, its impact is often limited when foundational biases within the training data remain unaddressed.

\textbf{Systemic and Long-term Improvements from Data Debiasing}. Data debiasing alleviate biases at their source by improving the quality and balance of the training data used to pre-train MLLMs. 
This approach excels in addressing systemic issues such as language imbalances and demographic biases, enabling MLLMs to perform more fairly across both high- and low-resource languages, as well as different demographic groups. This is demonstrated by improved accuracy in cross-lingual understanding \cite{x2,x5} and enhanced fairness in demographic sentiment analysis tasks \cite{g9,g12}.

Despite progress in addressing ethical concerns through debiasing, challenges still persist in achieving comprehensive and scalable solutions. 

\textbf{Trade-offs with Performance}.
Many debiasing techniques introduce trade-offs, where mitigating bias can result in reduced accuracy or fluency in downstream tasks \cite{y34,1201}. 
Balancing fairness and performance remains a critical concern, especially in low-resource languages, where available data is often scarce and the risk of model overfitting or bias amplification is higher. This challenge is a promising direction worth exploring in depth.

\textbf{Dependency on Context}.
The inherent diversity of language and culture makes achieving absolute fairness across all contexts difficult \cite{k4}. While constructing more balanced corpora is a viable solution for mitigating language bias, the effective collection and integration of high-quality, low-resource language texts remains a formidable challenge \cite{g4}, often requiring significant human, financial, and computational resources.

\textbf{Lack of Interpretability}. 
Existing methods for bias understanding and mitigation face a major challenge: the lack of interpretability. Biases may stem not only from the model itself but also from external factors like training data, algorithms, or task settings. Research into bias interpretability helps identify these sources, enabling more targeted debiasing strategies. Methods like bias attribution in neural networks \cite{y29} and identifying bias neurons \cite{1202} are crucial for improving fairness and transparency.

\section{Future Directions}\label{5}
This survey provides a holistic, systematic overview of the evolution of multilingual large language models. The MLLMs are still in a developing stage and thus there are still several challenges for future research, which we summarize below:

\begin{itemize}
    \item \textbf{Performance on Low-resource Languages}.
    MLLMs outperform monolingual LLMs in downstream tasks for high-resource languages, but their performance on low-resource languages remains unsatisfactory \cite{fd1}, which may be due to limited annotated data \cite{fd3} for low-resource languages and low lexical overlap between high-resource and low-resource languages \cite{fd2}.  Specializing MLLMs based on language families can be an efficient way to more easily share information across languages \cite{fd6}. In addition, how to find a more robust tokenizer for most languages is worth investigating as well.
    \item \textbf{Limited and Unbalanced Multilingual Corpora}.
    The performance of MLLMs largely depends on the training data’s quality, size, and diversity \cite{fd9}. However, there is only a limited amount of data available for most of the world's languages. The overwhelming English texts in corpora lead to MLLMs’ English-centric ability. 
    Even though for some high-resource languages where data is available, 
    previous work has shown that some commonly used multilingual resources have severe quality issues \cite{fd4}. How to collect much more high-quality, larger scale, and more diverse training data from various languages deserves further research.
    \item \textbf{Usage of Multimodal Data Sources}.
    Leveraging information from multimodal data sources such as speech and images can alleviate high reliance on text data. 
    Human cognition and perception capabilities rely on diverse information, 
    and the usage of multimodal data can better align with human intentions. 
    Supported by multimodal data equates to higher quality, more diverse training data. 
    However, how to achieve universal representation accurately by modality alignment poses a new challenge, deserving further investigation.
    \item \textbf{Evaluation of Multilingual LLMs}.
    The evaluation benchmarks for MLLMs are mainly based on the development of English task sets. However, these benchmarks are not fully applicable to other languages. Although some task sets can be translated into other languages, due to the differences between languages, the performance of the translated data set will be lower than the source language. Besides, current evaluation benchmarks are all task-centric, lacking a universal and flexible evaluation system. 
    The topic of how to collect high-quality multilingual evaluation datasets and build a system to properly evaluate 
    the true multilinguality of MLLMs is still undervalued.
    \item \textbf{Ethical Impact of Multilingual LLMs}. Multilingual LLMs can inherit biases present in their training data, leading to ethical risks of generation. Due to the high proportion of Western language data in training data, 
    the MLLMs are inclined to reflect Western-centric concepts \cite{fd8}. 
    How to mitigate biases and ensure fairness and cultural sensitivity in text generation are key challenges for the further development of MLLMs.

\end{itemize}




\bibliographystyle{fcs}
\bibliography{reference}

\begin{biography}{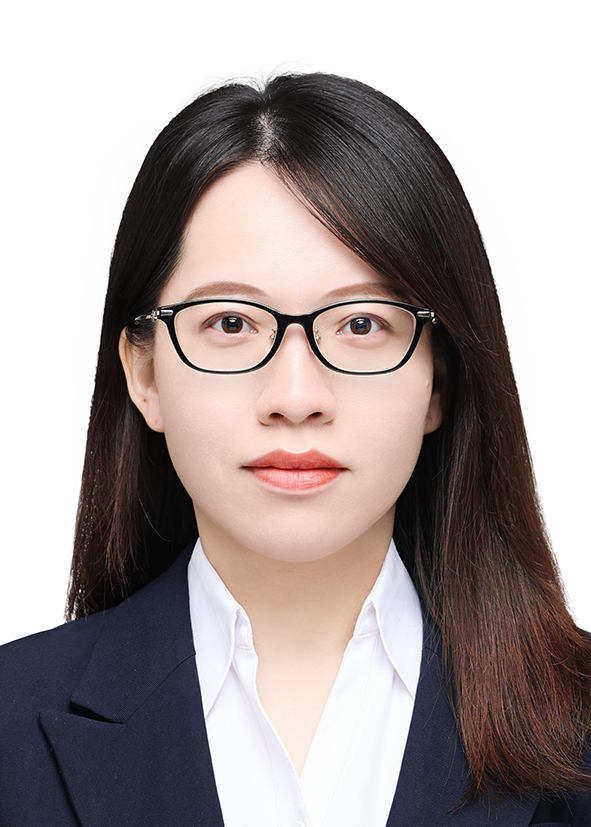}
Yuemei Xu is an associate professor in the School of Information Science and Technology, Beijing
Foreign Studies University. She received her PhD degree from Chinese
Academy of Sciences in 2014, the B.E. from Beijing University of Posts and
Telecommunications (China) in 2009. Her main research interests include
Multilingual Natural Language Processing and Artificial Intelligence.
\end{biography}

\begin{biography}{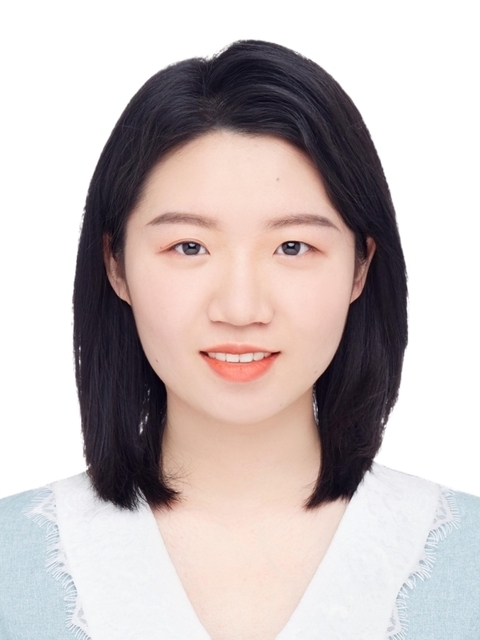}
Ling Hu received the bachelor's degree from Beijing University of Posts and
Telecommunications (China) in 2021. She is currently pursing the master degree with the School of Information Science and Technology, Beijing
Foreign Studies University.
Her main research interests include
Multilingual Natural Language Processing and Artificial Intelligence.
\end{biography}

\begin{biography}{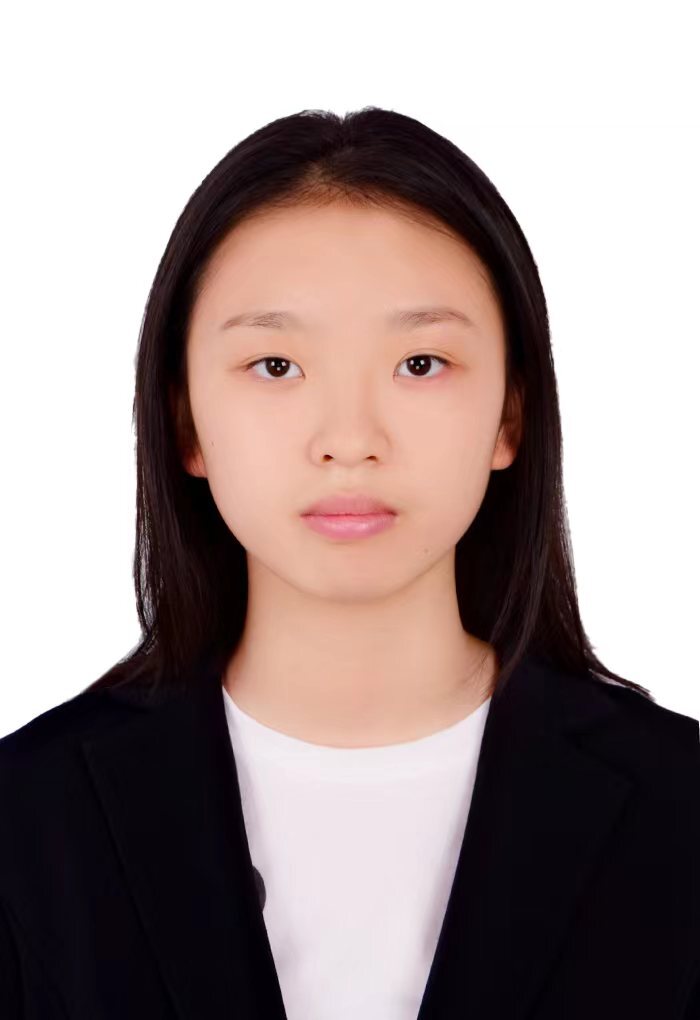}
Jiayi Zhao is majoring in computer science and technology at the School of Information Science and Technology, Beijing Foreign Studies University. Her main research interests include Multilingual Natural Language Processing and Artificial Intelligence.
\end{biography}

\begin{biography}{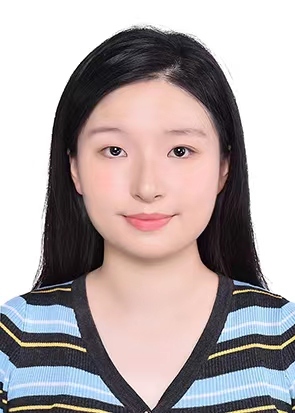}
Zihan Qiu is majoring in computer science and technology at the School of Information Science and Technology, Beijing Foreign Studies University. Her main research interests include Multilingual Natural Language Processing and Artificial Intelligence.
\end{biography}

\begin{biography}{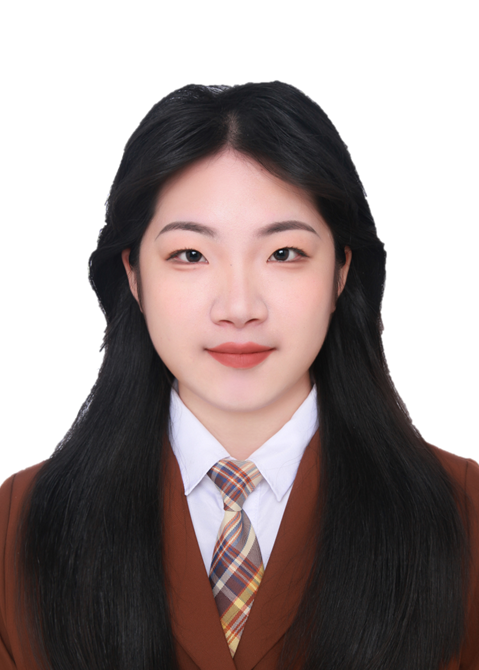}
Kexin Xu received the bachelor's degree from Southwestern University of Finance and Economics in 2024. She is currently pursing the master degree with the School of Information Science and Technology, Beijing Foreign Studies University.
Her main research interests include
Multilingual Natural Language Processing and Artificial Intelligence.
\end{biography}

\begin{biography}{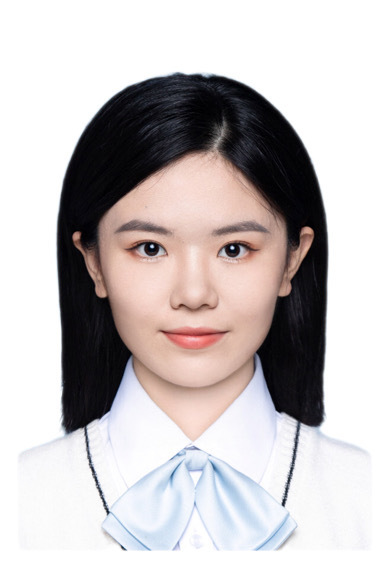}
Yuqi Ye is majoring in computer science and technology at the School of Information Science and Technology, Beijing Foreign Studies University. Her main research interests include Multilingual Natural Language Processing and Artificial Intelligence.
\end{biography}

\begin{biography}{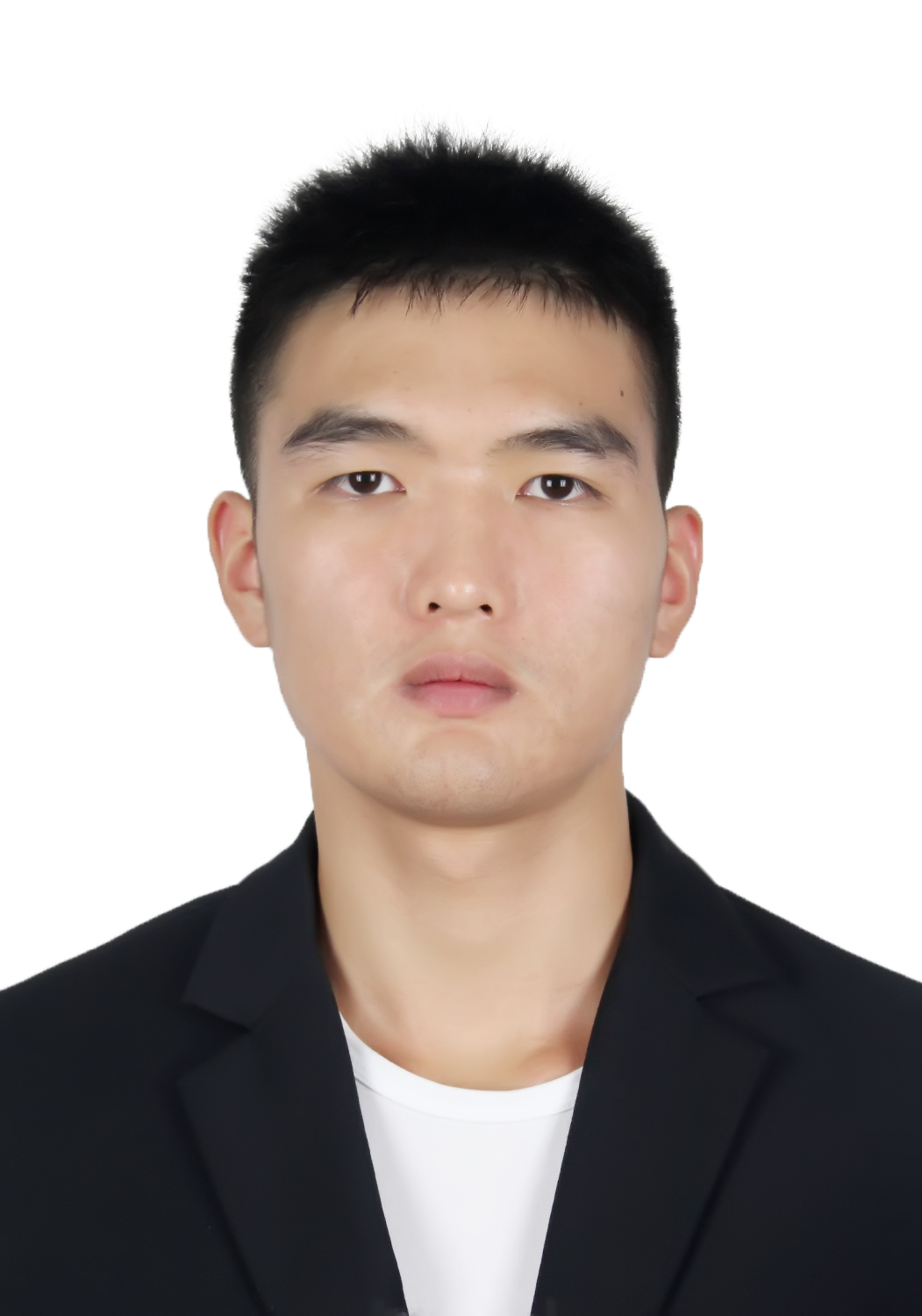}
Hanwen Gu received the bachelor of Engineering degree from the School of Information Science at Beijing Language and Culture  University in 2023. Currently, he is pursuing a master's degree in the School of Information Science and Technology at Beijing Foreign Studies University. His primary research interests encompass natural language processing and artificial intelligence.
\end{biography}

\end{CJK}
\end{sloppypar}
\end{document}